\documentclass{article}

     \usepackage[preprint]{nips_2018}

\usepackage[utf8]{inputenc} 
\usepackage[T1]{fontenc}    
\usepackage[]{hyperref}       
\usepackage{url}            
\usepackage{booktabs,tabularx}       
\usepackage[table]{xcolor}
\newcolumntype{Y}{>{\centering\arraybackslash}X}
\usepackage{amsfonts}       
\usepackage{nicefrac}       
\usepackage{microtype}      
\usepackage{graphicx}
\usepackage{multicol}
\usepackage{mathptmx}
\usepackage{amsmath}
\usepackage[export]{adjustbox}

\title{Thinking Aloud:\\ Dynamic Context Generation Improves Zero-Shot Reasoning Performance of GPT-2}

\author{%
  Gregor Betz
  \\
  Karlsruhe Institute of Technology\\
  Karlsruhe, Germany\\
  \texttt{gregor.betz@kit.edu} \\
   \And
  Kyle Richardson
  \\
  Allen Institute for AI\\ 
  Seattle, WA, USA\\
  \texttt{kyler@allenai.org} \\
   \And
  Christian Voigt
  \\
  Karlsruhe Institute of Technology\\
  Karlsruhe, Germany\\
  \texttt{christian.voigt@kit.edu} \\
}

\begin{document}

\maketitle

\begin{abstract}
  Thinking aloud is an effective meta-cognitive strategy human reasoners apply to solve difficult problems. We suggest to improve the reasoning ability of pre-trained neural language models in a similar way, namely by expanding a task's context with problem elaborations that are dynamically generated by the language model itself. Our main result is that dynamic problem elaboration significantly improves the zero-shot performance of GPT-2 in a deductive reasoning and natural language inference task: While the model uses a syntactic heuristic for predicting an answer, it is capable (to some degree) of generating reasoned additional context which facilitates the successful application of its heuristic. We explore different ways of generating elaborations, including fewshot learning, and find that their relative performance varies with the specific problem characteristics (such as problem difficulty). Moreover, the effectiveness of an elaboration can be explained in terms of the degree to which the elaboration semantically coheres with the corresponding problem. In particular, elaborations that are most faithful to the original problem description may boost accuracy by up to 24\%.
\end{abstract}


\section{Introduction}

Transformer-based language models \citep{Vaswani2017AttentionIA} have conquered, over the last three years, the leaderboards of NLP benchmarks -- bidirectional models like BERT \citep{Devlin2019BERTPO} and RoBERTa \citep{liu2019roberta} excel in ever more challenging natural language understanding (NLU) tasks, whereas autoregressive models such as BART \citep{lewis2019bart} or GPT-3 \citep{brown2020language} are capable of generating high quality texts that humans fail to tell apart from passages written by human authors \citep{brown2020language}. These technologies are not only reshaping the field of NLP, but are likely to have far-reaching repercussions for how we read, study, and write texts in academia (especially in the humanities and social sciences), and beyond. 

As language models are continuously improving in terms of language understanding and linguistic reasoning skill, the question that naturally arises is whether there are any upper limits on what these systems will be able to do (with words). Are there hard problems that language models will never master? Shane Frederic's cognitive reflection test, which includes the following question, is an interesting case in point \citep{Frederick:2005mt}:

\begin{quote}
In a lake, there is a patch of lily pads. Every day, the patch doubles in size. If it takes 48 days for the patch to cover the entire lake, how long would it take for the patch to cover half of the lake?
\end{quote}

Consider how a human reasoner might tackle such a question, assuming that the answer is not immediately clear. A skillful thinker would re-read, rephrase, and elaborate the problem, as well as develop and think through alternative solutions. She might do so silently, or aloud, before she provides, ultimately, her answer. Actually, thinking aloud has been empirically shown to improve problem solving capability and reading comprehension both in children \citep{gagne1962study,ahlum1986effect,SHORT1991109,silven1992improving} as well as in adults \citep{wetzstein2004reflective,Fox2010HowTG}. Moreover, such problem elaboration (individual or collaborative) is an established and well-studied teaching method \citep{lochhead1987teaching}. Now, is thinking aloud also a viable `meta-cognitive' strategy for language models as artificial reasoners? Can language models elaborate problems and does this help them to get the answers right? These are the questions we address in this study.

\begin{figure}
  \centering
  \includegraphics[width=\linewidth]{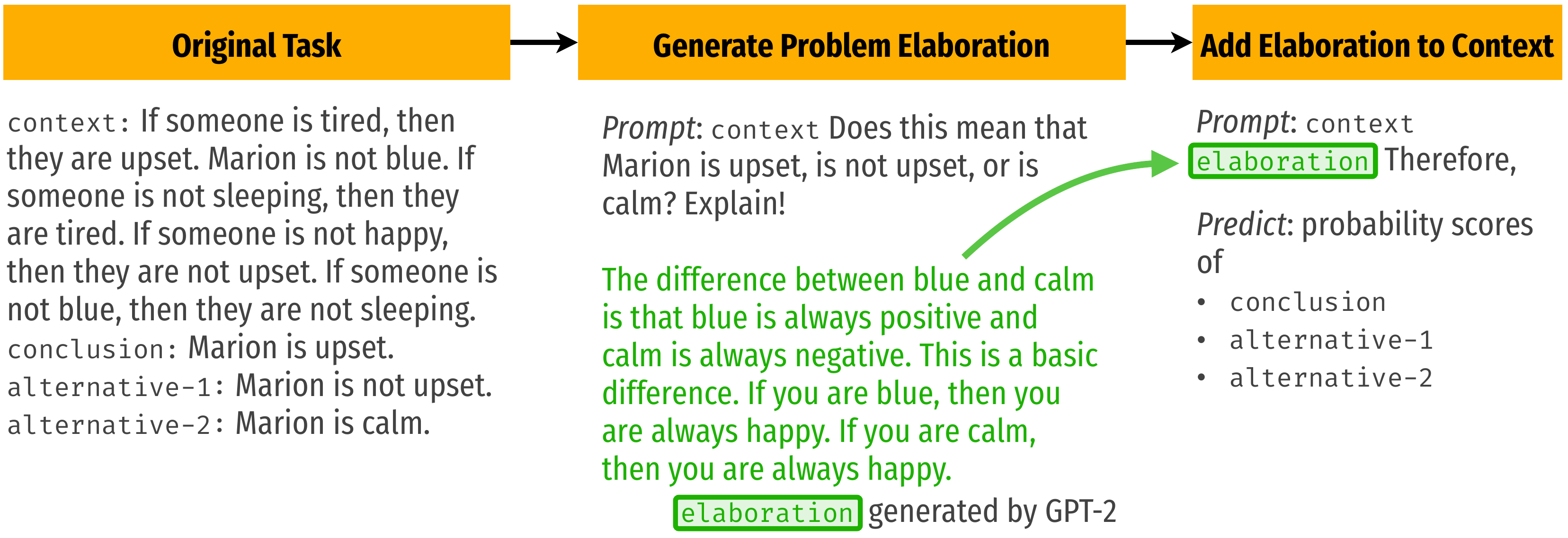}\\
  \caption{Dynamic problem elaboration of a reasoning task, illustrated by an example drawn from the ChainRuler dataset.}
  \label{fig:first_example}
\end{figure}

In the remainder of the paper, we will refer to the generation of text that effectively analyzes a given problem as "dynamic problem elaboration," rather than using the term "thinking aloud" (because of its mental presumptions). "Dynamic" means that the language model is supposed to newly generate the elaboration in response to being challenged, and specifically for the problem at hand. Moreover, we will investigate a bootstrapping scenario where one and the same language model is used to answer the question and to generate the problem elaboration. In other words, the language model expands the context of each problem and feeds it back to itself (see also Section\ \ref{subsection:contextretrieval}) before predicting the answer. The example in Figure\ \ref{fig:first_example} illustrates this basic idea.

We test the effect of different kinds of problem elaboration on the performance of GPT-2 \citep{Radford2019} in a deductive, multi-hop natural language reasoning task inspired by \citet{Clark2020_TransSoftReas} and named "ChainRuler" (Subsection~\ref{subsec:data}). Given a context consisting of natural language rules and facts (e.g., the context illustrated in Figure~\ref{fig:first_example}) the goal is to answer yes/no questions (e.g., \emph{Marion is upset?}) that, by construction, require performing correct deductive reasoning over the provided context (Subsection~\ref{subsec:predicting}). \emph{Free and fewshot elaborations} consist in text generated in response to a generic, unanswered question, whereas \emph{piecemeal elaborations} are assembled from multiple generated text fragments that address task-specific queries \citep[as, e.g., in][]{shwartz2020unsupervised} (Subsection~\ref{subsec:gen_elabs}). 

Here is a preview of our main results: GPT-2 follows a simple syntactic heuristic \citep[similiar to those discussed in][]{McCoy2019RightFT} when prompted with a ChainRuler reasoning problem, which, in benevolent cases, is effective and leads to high accuracy, but causes systematic bias as soon as there is sufficient \emph{effective distraction} or the task involves \textit{contraposition} (Section~\ref{sec:results}). Against this baseline, dynamic problem elaborations can, depending on the problem characteristics, increase accuracy by 9\% -- either improving zero-shot skill or effectively de-biasing the model (Section~\ref{sec:results}). The observed variance in elaboration effectiveness may be explained in view of the elaborations' coherence with the problem to be solved. Specifically, the most faithful piecemeal elaborations boost accuracy by 24\% resp.\ 16\%, compared to the no elaboration treatment (Subsection~\ref{subsec:verisim_pert_faithf}). Likewise, different types of fewshot elaborations excel under different kinds of problem characteristics (Section~\ref{sec:results}) -- especially so when negations are absent in the corresponding problem descriptions (Subsection~\ref{subsec:similsamplesolu}).

\section{Related Work}

\subsection{Reasoning Performance of Neural Language Models}

Most of the studies that assess reasoning and inference skills of neural language models (LMs) seem to support the following claims \citep[see also][]{Rogers2020API}:

\emph{1. Pre-trained neural language models, whether uni- or bidirectional, display a poor zero-shot per\-formance on reasoning tasks} \citep[e.g.,][]{yanaka2019can, Clark2020_TransSoftReas, Richardson:2020ym}. Even GPT-3, while achieving impressive zero-shot results for other NLU benchmarks, struggles with the task of natural language inference (NLI) in particular \citep[Sec.\ 3.8]{brown2020language}. Moreover, \citet{kassner2020negated}, extending the LAMA probe by \citet{Petroni:2020ih}, show that LMs are vulnerable to mispriming effects and have major difficulties in getting negations right \citep[consistent with][]{talmor2020olmpics}. Similarly, \citet{Richardson:2020ym} probe language models with semantic fragments and find that even models that are fine-tuned on NLI datasets fail to cope with, e.g., elementary logical relations. However, there is evidence that pre-trained language models do possess substantial \emph{conceptual} knowledge, which shows in their ability to correctly draw conceptual (as opposed to formal or logical) inferences \citep{Richardson2019WhatDM,talmor2020olmpics} and to rely on these relations as implicit background assumptions in answering questions \citep{talmor2020leap}.

\emph{2. With task-specific fine-tuning or so-called inoculation \citep{liu2019inoculation}, however, these models can achieve state-of-the art results and are almost perfectly mastering many reasoning tasks.} While zero-shot performance is generally poor, language models trained on task-specific data have propelled SOTA accuracy levels above 90\% for major benchmarks (such as SNLI \citep{snli:emnlp2015}, MultiNLI \citep{Williams2018ABC_MultiNLI} and RTE \citep{Dagan2005ThePR}). Language models quickly learn to master logical fragments given appropriate training data \citep{kassner2020negated, Richardson2019WhatDM, Richardson:2020ym}, and can be fine-tuned to correctly draw complex deductive inferences \citep{Clark2020_TransSoftReas,betz2020critical} and to generate informal reasons \citep{rudinger-etal-2020-thinking,camburu2018snli,brahman2020learning}.
\citet{schick2020exploiting,schick2020its} introduce "Pattern Exploiting Training" (PET) and show that unsupervised pattern-recognition and annotation of training data substantially boosts the performance of the language model that is trained on the labeled data.

Against this background, the novelty of our study is to show that GPT-2 has a strong zero-shot performance on a NLI task involving deductive reasoning over rules \citep{Clark2020_TransSoftReas}. Our particular focus on zero-shot performance follows much recent work on zero-shot evaluation of pre-trained language models \citep{shwartz2020unsupervised,ma2020knowledge,banerjee2020self,bosselut2019dynamic}, which take zero-shot performance of pre-trained models without specialized fine-tuning as an insightful benchmark for better understanding LM's reasoning abilities.

\subsection{Dynamic Templating and Context Retrieval}\label{subsection:contextretrieval}

It is well-known that performance of neural LMs in NLP tasks depends sensitively on the wording of the query \citep{Petroni:2020ih,Jiang2020}. Accordingly, \citet{Petroni:2020ih} argue that by assessing a language model with a given set of manually defined queries one measures a lower bound of the system's full skill. Recent studies have explored two directions for dynamically adjusting and expanding LM queries, which are conceptually related to automatic query expansion in (classic) information retrieval systems \citep{Carpineto2012}:

\emph{1. Dynamic templating refers to the automatic and case-specific optimization of natural language templates which are used to construct a query from given data.} Specifically, \cite{Jiang2020} explore three strategies for improving manually generated prompts: mining effective prompts from a database (e.g., Wikipedia), paraphrasing prompts (e.g., through two-way-translation, forth and back), and pooling multiple prompts. Each of these strategies is shown to significantly improve predictions in QA tasks.     

\emph{2. Dynamic context expansion refers to the automatic retrieval and/or generation of additional context, over and above the task data, that is embedded in the query.} \citet{Chen2019MultihopQA} extract and add "reasoning chains" to problem descriptions, which improves performance on multi-hop QA tasks \citep{Yang2018HotpotQAAD}. Likewise,  \citet{Petroni2020HowCA} assess whether automatic context expansion boosts the performance of the RoBERTa model \citep{liu2019roberta} on a QA task. Standard information retrieval systems are shown to increase accuracy by 10\% to 40\%, depending on the specific task. If, however, a text that is generated with a language model is added to the context, precision actually drops as compared to the baseline performance without context expansion \citep{Petroni2020HowCA}. Whereas such free context expansion deteriorates performance, \citet{shwartz2020unsupervised}, introducing \emph{self-talk}, demonstrate that task-specific and highly structured generation of additional context (called "clarifications") may improve the performance of various language models in commonsense QA tasks. Retrieval augmented generation (RAG) \citep{Lewis2020RetrievalAugmentedGF} pushes dynamic context expansion one step further by coupling, in one global net, a transformer-based neural document retrieval system with a language model for answer prediction. RAG leads to substantially more accurate, factive and specific answers than obtained by the bare generative language model \citep{Lewis2020RetrievalAugmentedGF}. Moreover, dynamic context expansion has recently been successfully applied to reasoning tasks. PRover \citep{saha2020prover} is a multi-head model based on RoBERTa \citep{liu2019roberta}, which both constructs proof chains and predicts answers for a deductive reasoning task \citep{Clark2020_TransSoftReas}. \citet{saha2020prover} show that this kind of structured problem elaboration significantly boosts accuracy (by 6\% for zero-shot setting). Likewise, \citet{gontier2020measuring} demonstrate that transformer language models can be trained to generate effective context expansions that allow the model to solve reasoning tasks from CLUTTR, a database of inference problems with implicit general premises \citep{Sinha:2019va}.       

Against this background, the novelty of our study consists in showing that bootstrapping context generation, where one and the same language model that is used for answer prediction is also employed for dynamic context expansion, can increase the zero-shot reasoning performance of an autoregressive transformer model in a NLI task.

\section{Experiments}

We study the effect of problem elaboration on GPT-2's reasoning skill by adding different types of dynamically generated texts to the context in a reasoning task. Roughly speaking, we proceed in three steps: First, we synthesize test data for our reasoning task (Subsection~\ref{subsec:data}). Second, we generate and add problem elaborations for each entry in the dataset  (Subsection~\ref{subsec:gen_elabs}). Third, we append the generated elaborations to the context and predict the answers (Subsection~\ref{subsec:predicting}).

\subsection{Synthesizing the ChainRuler Data}\label{subsec:data}

In order to test the zero-shot reasoning skill of GPT-2 and the effectiveness of dynamic problem elaboration, we design a deductive reasoning task, inspired by RuleTaker \citep{Clark2020_TransSoftReas}, and construct a corresponding synthetic dataset. In a nutshell, the task consists in correctly inferring a conclusion from a set of rules and a fact. More specifically, each problem is composed of:

\begin{enumerate}
    \item the \textbf{conclusion} (correct answer): a singular, possibly negated statement (e.g., "$a \textrm{ is } G$"); 
    \item two \textbf{false alternatives} which contradict the conclusion: the logical negation of the conclusion ("$a \textrm{ is not } G$") and a singular statement which contradicts the conclusion for conceptual reasons ("$a \textrm{ is } \bar{G}$" with $\bar{G}$ being conceptually complementary to $G$);
    \item the \textbf{fact}: an singular statement "$a \textrm{ is } F$" (or, "$a \textrm{ is not } F$"), which serves as premise;
    \item the \textbf{rule chain}:  $l$ generalized conditionals that allow one to infer the correct answer from the \emph{fact} ($F \supset I_{1}$,  $I_{1} \supset I_{2}$, \ldots,  $I_{l-1} \supset G$). If the problem is of type "\emph{contraposition}", then the last conditional is transposed (replaced by $\textrm{not-}G \supset \textrm{not-}I_{l-1}$);
    \item the \textbf{distractors}: a set of $k$ confounding rules whose consequent terms equal the target predicate or its logical / conceptual complement: $H_1 \supset X_{1}$,  $H_{2} \supset X_{2}$, \ldots,  $H_{k} \supset X_k$ with $X_i \in \{G, \textrm{not-}G, \bar{G}, \textrm{not-}\bar{G}\}$.  
\end{enumerate}

The \emph{problem description} (\texttt{context}) of a single task item consists in a random permutation of the fact, the relevant rules (rule chain) and the confounding rules (distractors). By "\emph{depth}" of a problem, we refer to the length $l$ of its rule chain, whereas the \emph{breadth} denotes the number $k$ of confounding rules.

Note that the distractors, and especially their consequent terms, are sampled randomly. So, by mere chance, all confounding rules in a problem description might actually point towards the correct answer (consequent term = target predicate). To capture this property of a problem, we introduce the notion of \emph{effective distraction}, which is the number of distractors whose consequent term is not identical with the conclusion's predicate.

\begin{table}[tbp]
  \caption{ChainRuler task examples} 
  \label{table:chainrulerexample}
  \begin{small}
  \begin{tabular}{@{}p{0.25\linewidth}@{}p{0.75\linewidth}@{}}
    \toprule
    \texttt{fact} & "Jill is green." \\
    \texttt{rulechain} & "If someone is green, then they are loud.", "If someone is loud, then they are guilty."\\
    \texttt{distractors} & "If someone is empty, then they are innocent." \\
    \texttt{conclusion} & "Jill is guilty." \\
    \texttt{alternatives} & "Jill is not guilty.", "Jill is innocent." \\
    \texttt{depth} & 2 \\
    \texttt{breadth} & 1 \\
    \texttt{contraposition} & False \\
    \texttt{eff.\ distraction} & 1 \\
    \midrule
    \texttt{fact} & "Lily is blue." \\
    \texttt{rulechain} & "If someone is blue, then they are careful.", "If someone is careful, then they are loud.", "If someone is not generous, then they are not loud."\\
    \texttt{distractors} & "If someone is in need of money, then they are not generous.", "If someone is guilty, then they are not generous." \\
    \texttt{conclusion} & "Lily is generous." \\
    \texttt{alternatives} & "Lily is not generous.", "Lily is stingy." \\
    \texttt{depth} & 3 \\
    \texttt{breadth} & 2 \\
    \texttt{contraposition} & True \\
    \texttt{eff.\ distraction} & 2 \\
    \bottomrule
  \end{tabular}
  \end{small}
\end{table}

The construction of the synthetic test dataset can be broken down in two main steps:

\begin{enumerate}
    \item We randomly sample a balanced set of formal problem descriptions that instantiate the above structure, while systematically varying problem characteristics such as depth and breadth.   
    \item Drawing from a database of (i) names, (ii) pairs of conceptually contradictory predicates, and (iii) simple natural language templates, we create natural language instances of the formal problems by simple substitution.     
\end{enumerate}

Table~\ref{table:chainrulerexample} illustrates the ChainRuler task by presenting two example items from the dataset.

\subsection{Generating Problem Elaborations}\label{subsec:gen_elabs}

\begin{figure}
  \centering
  \includegraphics[width=\linewidth]{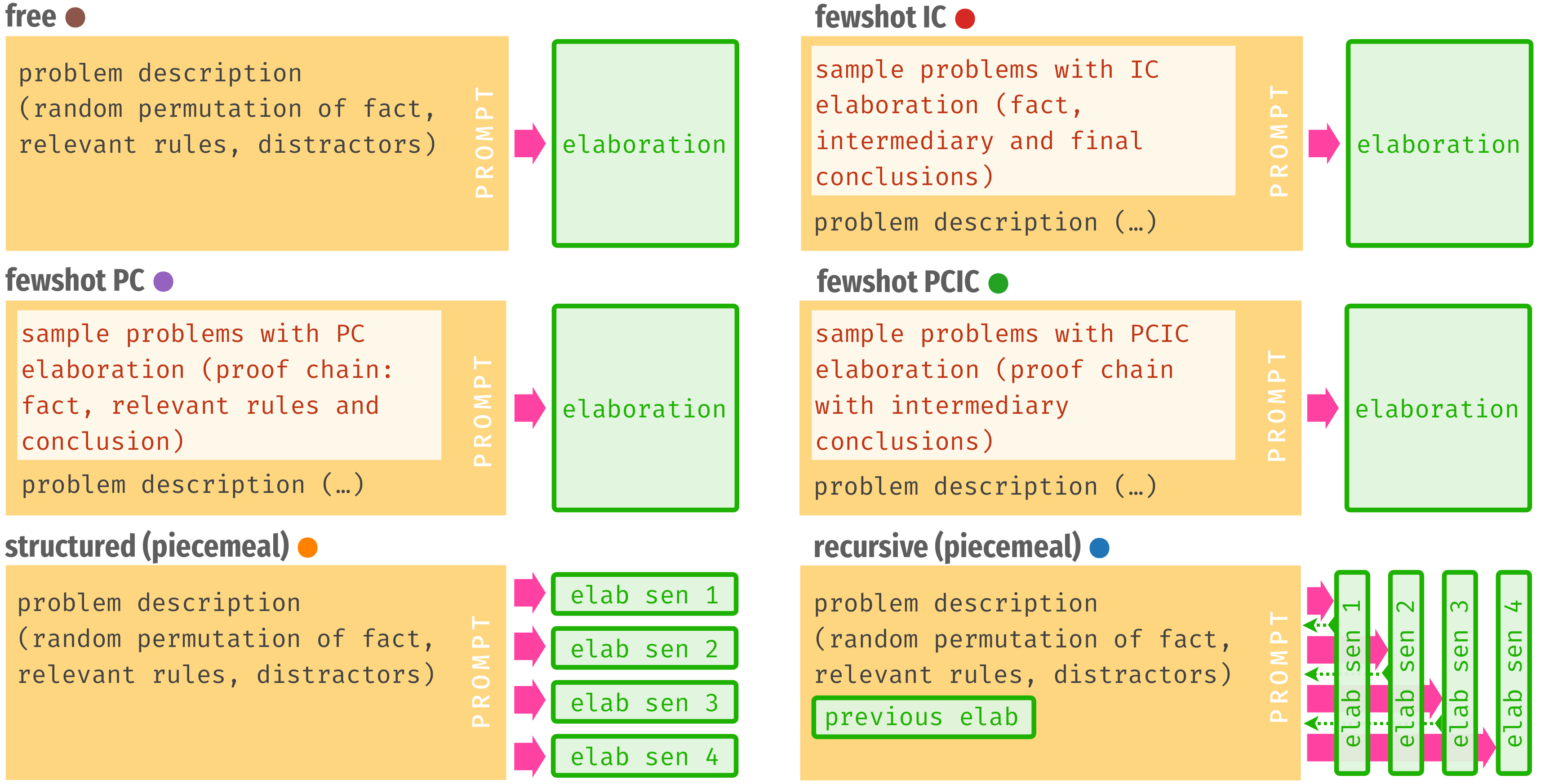}
  \caption{Illustration of the different methods for eliciting and generating problem elaborations studied in this paper.}
  \label{fig:illustr_elabs}
\end{figure}

We distinguish and study six ways of generating problem elaborations (cf.\ Figure~\ref{fig:illustr_elabs}).

\textit{Free elaboration}. We prompt the model with an unanswered question and generate one single completion. The first four sentences of this generated completion represent the "free elaboration" of the problem. The query for eliciting this free elaboration presents the \texttt{context} and asks which of the alternative answers is correct, e.g.: "Here is what we know: \texttt{context} Does this mean that Loretta is not hungry, is hungry, or is not full? Explain!"

The fewshot elaborations are generated similarly to the free elaborations, with the exception that two "sample solutions" are prepended to the prompt. Each sample solution features a problem description and a proof chain serving as paradigmatic elaboration of the problem. More specifically, we explore the following three kinds of paradigmatic elaborations:  

\begin{itemize}
    \item \emph{IC elaboration}, which consists in the problem's fact, the intermediary conclusions that can be inferred from the fact by consecutively applying the rules, and the final conclusion;
    \item \emph{PC elaboration}, which consists in the problem's fact, the rule chain (correctly ordered), and the final conclusion;
    \item \emph{PCIC elaboration}, which consists in the problem's fact, followed alternately by the relevant rules and conclusions one can infer, until the final conclusion is reached. 
\end{itemize}

This gives, correspondingly, the following fewshot elaborations: \textit{Fewshot IC}, \textit{Fewshot PC}, and \textit{Fewshot PCIC}. 

With free and fewshot elaboration, the model generates, given its prompt, a single completion. Structured and recursive elaboration, described below, are, in contrast, \emph{piecemeal} methods, which prompt the model not once but four times. The four generated completions are then post-processed and concatenated to obtain the problem elaboration.

\textit{Structured elaboration}. The model generates, independently of each other, four completions given one and the same prompt. The four sentences which come first in each conditionally generated text are concatenated and represent the "structured elaboration" of the problem. The specific query used to elicit the structured elaboration states the \texttt{context} and ends with a cloze-style question about what one may infer about the \emph{subject}, e.g.: "Here is what we know: \texttt{context} Therefore, Loretta".

\textit{Recursive elaboration}. The model generates a single sentence given the prompt used for structured elaboration. The generated sentence is then added to the context, before the model is prompted again to generate a second sentence, which is once more appended to the context, and so on, until four sentences are iteratively generated. These four statements make up the recursive elaboration of the problem. 

\begin{table}[htbp]
  \caption{Elaboration examples, corresponding to the entries in Table\ \ref{table:chainrulerexample}. We color generated sentences in accordance with their logical relation to the given \texttt{context} (\textcolor{blue!60!darkgray}{independent/digression}, \textcolor{green!30!darkgray}{implied/explicit in \texttt{context}}, \textcolor{green!80!red}{implied/implicit in \texttt{context}}, \textcolor{red!80!darkgray}{inconsistent}).}
  \label{table:elaboration_examples}
  \begin{small}
  \begin{tabular}{@{}p{0.1\linewidth}@{\rule[0mm]{0.05\linewidth}{0cm}}p{0.85\linewidth}@{}}
    \toprule
    \texttt{context} & \textit{If someone is empty, then they are innocent. If someone is green, then they are loud. If someone is loud, then they are guilty. Jill is green.} \\
    \texttt{free} & \textit{Does this mean that Lily is generous, is not generous, or is stingy? Explain!} \textcolor{blue!60!darkgray}{5.2.3: What is the difference between innocent and guilty?  If you are guilty, you are not innocent.}\\
    \texttt{fewshot IC} & \textit{Does this mean that Lily is generous, is not generous, or is stingy? Explain!} \textcolor{green!30!darkgray}{Well, it says that Jill is green.} \textcolor{red!80!darkgray}{It follows that Jill is innocent. Therefore, Jill is innocent.}\\
    \texttt{fewshot PC} & \textit{Does this mean that Lily is generous, is not generous, or is stingy? Explain!} \textcolor{green!30!darkgray}{Well, it says that Jill is green. If someone is green, then they are loud. If someone is loud, then they are guilty.} \textcolor{green!80!red}{Therefore, Jill is guilty.}\\
    \texttt{fewshot PCIC} & \textit{Does this mean that Lily is generous, is not generous, or is stingy? Explain!} \textcolor{green!30!darkgray}{Well, it says that Jill is green. If someone is green, then they are loud.} \textcolor{green!80!red}{It follows that Jill is loud.} \textcolor{red!80!darkgray}{If someone is loud, then they are innocent.} \textcolor{red!80!darkgray}{Therefore, Jill is innocent.} \\
    \texttt{structured} & \textit{Therefore, we may conclude that Jill} \textcolor{green!80!red}{is guilty.} \textit{Jill} \textcolor{green!80!red}{is guilty.} \textit{Jill} \textcolor{green!80!red}{is guilty.} \textit{Jill} \textcolor{green!80!red}{is guilty.} \\
    \texttt{recursive} & \textit{Therefore, we may conclude that Jill} \textcolor{green!80!red}{is guilty.} \textit{Jill} \textcolor{green!80!red}{is guilty.} \textit{Jill} \textcolor{green!80!red}{is guilty.} \textit{Jill} \textcolor{green!80!red}{is guilty.}\\
    \midrule
    \texttt{context} & \textit{If someone is not generous, then they are not loud. If someone is blue, then they are careful. Lily is blue. If someone is in need of money, then they are not generous. If someone is guilty, then they are not generous. If someone is careful, then they are loud.} \\
    \texttt{free} & \textit{Does this mean that Lily is generous, is not generous, or is stingy? Explain!} \textcolor{blue!60!darkgray}{You may have already seen the next three sentences: "If someone is not generous, then they are not loud." "If someone is not blue, then they are careful." "If someone is in need of...}\\
    \texttt{fewshot IC} & \textit{Does this mean that Lily is generous, is not generous, or is stingy? Explain!} \textcolor{green!30!darkgray}{Well, it says that Lily is blue.} \textcolor{green!80!red}{It follows that Lily is not in need of money. Therefore, Lily is not in need of money.}\\
    \texttt{fewshot PC} & \textit{Does this mean that Lily is generous, is not generous, or is stingy? Explain!} \textcolor{green!30!darkgray}{Well, it says that Lily is blue.} \textcolor{green!80!red}{If someone is in need of money, then they are not careful. If someone is guilty, then they are not careful.} \textcolor{green!30!darkgray}{If someone is careful, then they are loud.} \textcolor{red!80!darkgray}{Therefore, Lily is not generous.}\\
    \texttt{fewshot PCIC} & \textit{Does this mean that Lily is generous, is not generous, or is stingy? Explain!} \textcolor{green!30!darkgray}{Well, it says that Lily is blue. If someone is blue, then they are careful.} \textcolor{green!80!red}{It follows that Lily is careful.} \textcolor{green!30!darkgray}{If someone is careful, then they are loud.} \textcolor{green!80!red}{And therefore, Lily is loud.} \\
    \texttt{structured} & \textit{Therefore, we may conclude that Lily} \textcolor{red!80!darkgray}{is not a generous person.} \textit{Lily} \textcolor{blue!60!darkgray}{is a friend of our kind.} \textit{Lily} is a blue. \textit{Lily} \textcolor{red!80!darkgray}{is not a blue person.} \\
    \texttt{recursive} & \textit{Therefore, we may conclude that Lily} \textcolor{red!80!darkgray}{is in need of money.} \textit{Lily} \textcolor{red!80!darkgray}{is in need of money.} \textit{Lily} \textcolor{green!80!red}{is not in need of money.} \textit{Lily} \textcolor{green!80!red}{is not in need of money.}\\
    \bottomrule
  \end{tabular}
  \end{small}
\end{table}

The free and structured elaborations are generated with top-p nucleus sampling (we follow \cite{shwartz2020unsupervised} in setting $p=0.5$). The remaining elaborations are decoded with beam search. Table\ \ref{table:elaboration_examples} displays examples of thusly elicited elaborations for two different ChainRuler problem items. To put the results in perspective, we compare the effects of dynamic problem elaboration with four synthetic context expansions that can be directly generated from the test data as follows:

\emph{Answers (Baseline):} We randomly pick one of the three alternative answers and repeat it four times.

\emph{Context (Baseline):} We concatenate four randomly picked statements from the \texttt{context}.  

\emph{Intermediary conclusions (Oracle):} We adjoin all intermediary conclusion about the \texttt{subject} that can be inferred by successively applying the given \texttt{rules} to the \texttt{fact}.  

\emph{Final conclusion (Oracle):} We repeat the final \texttt{conclusion} (i.e., the correct answer) four times.

\subsection{Predicting Answers}\label{subsec:predicting}

To predict answers, we calculate the conditional probability that the language model assigns to each possible answer given the \texttt{context} and -- depending on the experimental treatment -- the corresponding \texttt{elaboration}. The most likely answer is then predicted to be the correct one. Formally, consider context $c$, elaboration $e$ and possible answers $a_1,a_2,a_3$. Let $\mathrm{p}(s|s_c)$ be the conditional probability our language model assigns to sequence $s$ given sequence $s_c$ (as prompt). The correct answer is predicted according to ${\mathrm{argmax}_{i=1,2,3}}\big ( \mathrm{p}(a_i|c,e) \big ).$

In order to assess the quality of the model's probabilistic predictions, we reduce the problem to a binary classification task, where for each context $c$ and elaboration $e$ either $a$ or not $\lnot a$ is the correct answer. (We drop, in effect, the second false alternative from each item's answer set, cf.\ Section~\ref{subsec:data}.) The probabilistic scores for this binary task are obtained by normalizing the corresponding conditional probabilities, e.g. $\mathrm{prob}(a) = {\mathrm{p}(a|c,e)}/\big[{\mathrm{p}(a|c,e)+\mathrm{p}(\lnot a|c,e)}\big]$ and likewise for $\lnot a$, so that $\mathrm{prob}(a) + \mathrm{prob}(\lnot a)=1$.

Throughout this study, we use the HuggingFace implemention of the 1.5B-parameter GPT-2 model \citep{wolf2019huggingface}. 

As should be transparent from this Section's outline of the experiment, our study does not involve any training. We merely assess the zero-shot performance of the pre-trained GPT-2 model.

\section{Results}\label{sec:results}


\begin{figure}
  \centering\footnotesize
  \begin{tabular}{@{}c@{}c@{}c@{}c@{}c@{}c@{}}
    \multicolumn{2}{c}{(a) accuracy(\emph{none})} & \multicolumn{2}{c}{(b) $\Delta$acc(\emph{best\_elab,none})} & \multicolumn{2}{c}{(c) \emph{best\_elab}}  \\ \cmidrule(r){1-2} \cmidrule(r){3-4} \cmidrule(r){5-6}
    \emph{no cntrp} & \emph{cntrp} &
    \emph{no cntrp} & \emph{cntrp} &
    \emph{no cntrp} & \emph{cntrp} \\
    \includegraphics[height=22mm]{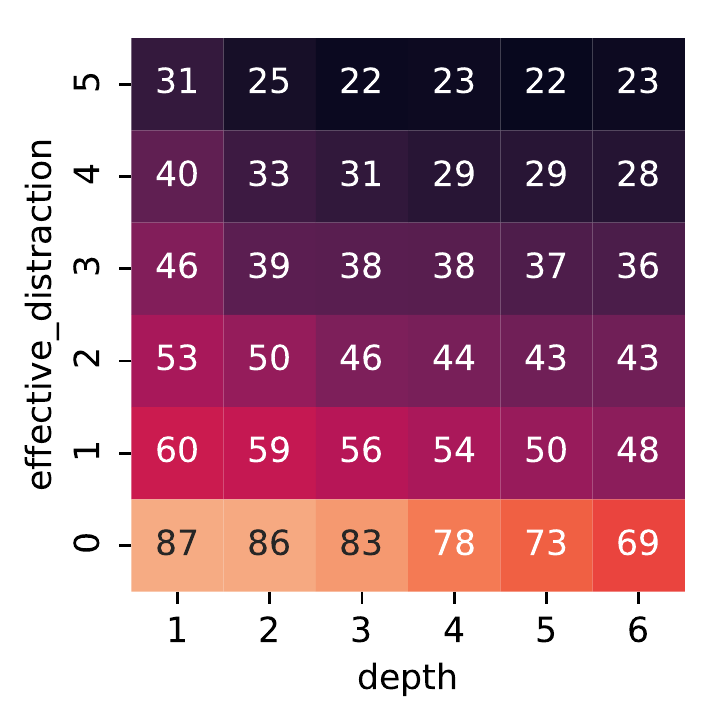} &  
    \includegraphics[height=22mm]{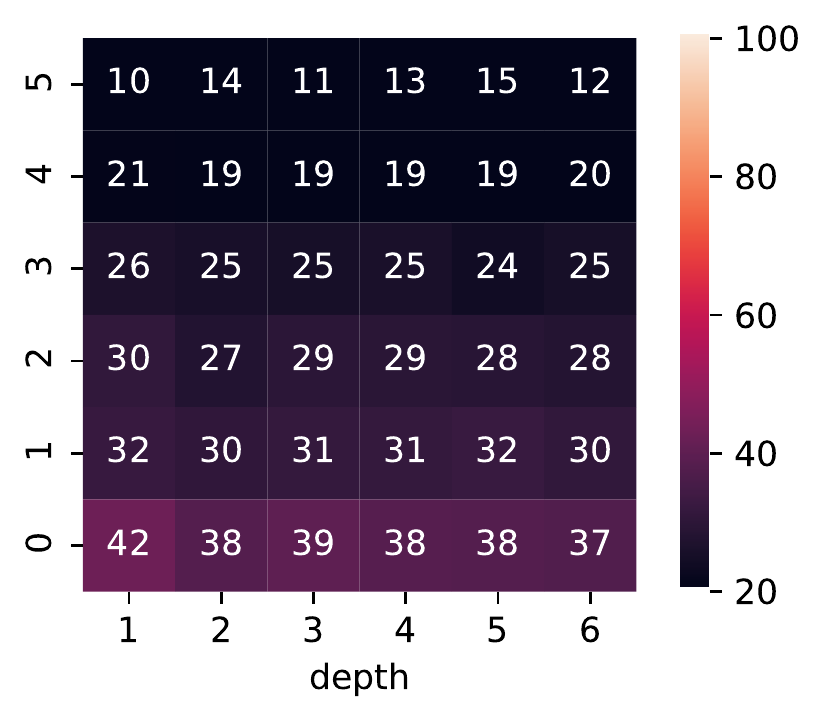}& 
    \includegraphics[height=22mm]{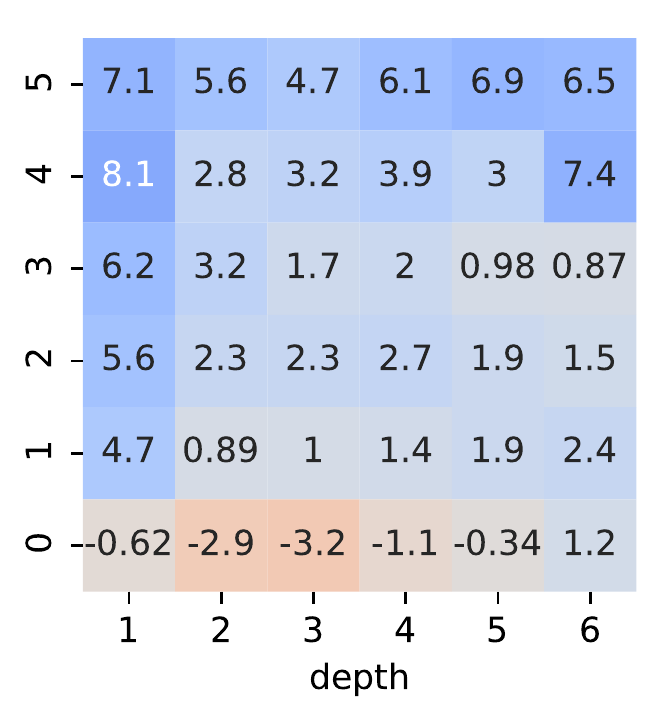} &
    \includegraphics[height=22mm]{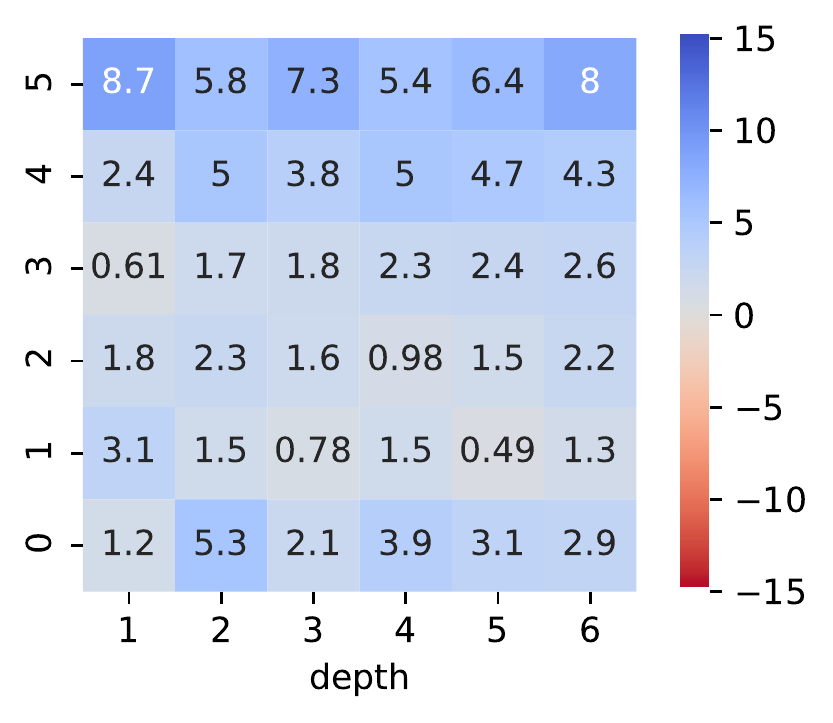} &
    \includegraphics[height=22mm]{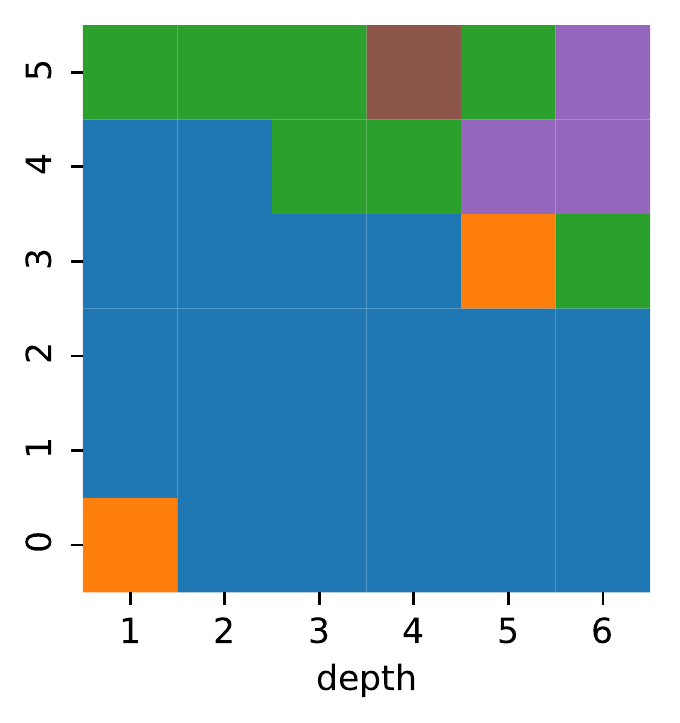}&
    \includegraphics[height=22mm]{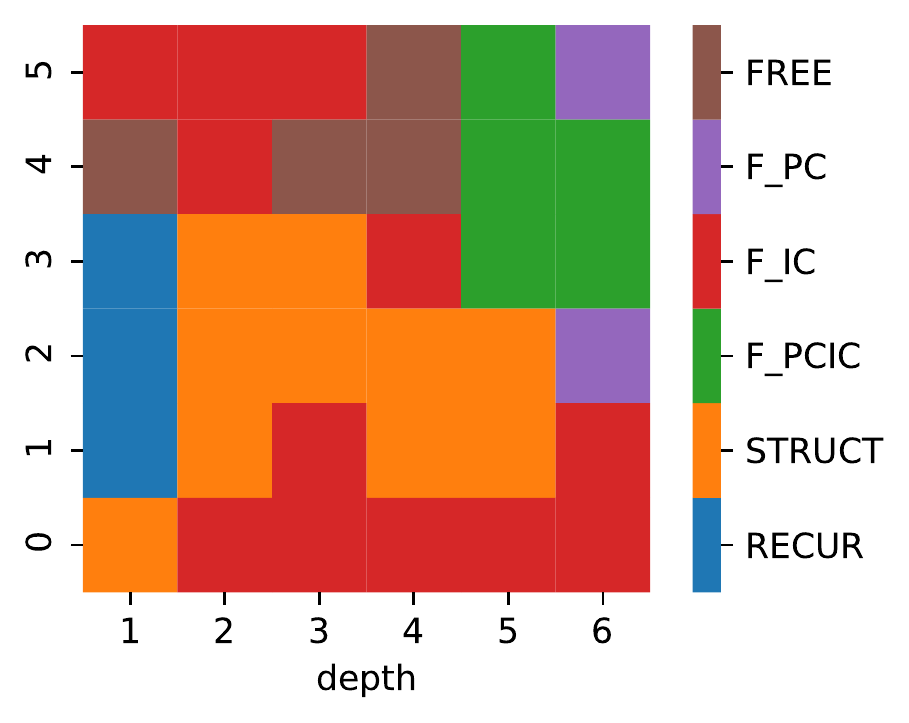}\\
  \end{tabular}
  
  \caption{Accuracy in ChainRuler tasks of given effective distraction and depth.  Subplots (a): absolute accuracy without elaboration (baseline \emph{none}). Subplots (b): relative accuracy gains for best-performing elaboration compared to baseline \emph{none}. Subplots (c): name of best-performing elaboration.}
  \label{fig:1st-o-acc}
\end{figure}

First of all, we find that GPT-2 follows a \textbf{simple heuristic} for solving task ChainRuler task: it's predictions are seemingly just based on how frequently the predicate of an answer-option appears in the consequent of the problem's rules. Whenever a problem description contains, by chance, many distractors whose "then"-part corresponds to the correct answer, GPT-2 achieves very high accuracy. This can be seen from Figure~\ref{fig:1st-o-acc}a, which displays \emph{no elaboration} accuracy as a function of a problem's \emph{depth} and its \emph{effective distraction} (see Section~\ref{subsec:data}). If the model is, however, not lucky and many distractors conincidentally point towards the wrong answers (i.e., high effective distraction), then the model typically gets the answer wrong and performs substantially worse than na\"ive random guessing (accuracy=.33). Following the simple heuristic, the model systematically commits fallacies and is substantially biased. This is especially the case in tasks \emph{with} contraposition, where the simple heuristic doesn't even work in the absence of distractors and performance is, accordingly, particularly weak. All this suggests that the pre-trained model does \emph{not} consecutively apply modus ponens, modus tollens, or chain rule to \emph{infer} the correct answer. It does \emph{not}, per se, engage in deductive reasoning. To further corroborate this conjecture, we have, in an additional experiment, replaced all antecedent conditions in the problems' rules with unrelated / non-sense statements, which, however, doesn't affect GPT-2's zero-shot performance on the tasks.  

Against this background, dynamic problem elaborations have a twofold effect (Figure~\ref{fig:1st-o-acc}b): On the one hand, they prevent the model from effectively applying its simple heuristic and hence reduce performance in cases the model was lucky and baseline accuracy was high (esp. \emph{no cntrp} and \emph{effective distraction}=0). On the other hand, dynamic problem elaboration both is a successful \textbf{de-biasing strategy}, and can further \textbf{boost reasoning performance}. If the baseline performance is worse than random guessing (e.g., if \emph{effective distraction}>3), then context expansion increases accuracy by up to 9 percentage points. In cases with slight distraction (e.g., \emph{no cntrp}, \emph{effective distraction}=2), the substantial baseline performance is further increased by up to 6 percentage points. All in all, the observed performance gains are in the upper range reported by \citet{shwartz2020unsupervised} for the similar \emph{self-talk} design in commonsense QA tasks.

Moreover, there is no single type of dynamic elaboration which performs best across the entire spectrum of different tasks (Figure~\ref{fig:1st-o-acc}c, see also Appendix~\ref{sec:app_elabmethods}): 
Without contraposition, recursive elaboration performs mostly best (and actually outperforms the \emph{intermediary conclusions oracle elaboration}) unless effective distraction and depth are very high. In the latter, arguably most difficult cases, fewshot elaborations are most effective. Regarding problems with contraposition, fewshot IC elaboration is top in problems with few effective distractors or many distractors and low depth; fewshot elaborations with proof chains are efficient given high effective distraction and great depth; and piecemeal elaborations perform best otherwise. One emerging overall pattern here is that fewshot elaborations tend to perform better than piecemeal elaborations if the task is very difficult and the model is negatively biased (baseline below random guessing).

\section{Analysis and Discussion}\label{sec:analysis}

The findings so far can be summarized as follows. GPT-2 follows a simple heuristic and predicts answers in line with their frequency of previous occurrence when prompted with a ChainRuler problem. This heuristic is effective in some lucky cases, but quickly leads to systematic failure when effective distraction increases. While dynamic problem elaboration decreases the effectiveness of the heuristic in the lucky cases, it also reduces bias and substantially improves performance across a wide-spectrum of problem constellations. Different elaboration methods display characteristic performance fingerprints.

In the following, we further differentiate and explain these results in terms of

\begin{enumerate}
    \item the degree to which generated elaborations facilitate the successful application of the simple syntactic heuristic used by the model (Subsection~\ref{subsec:luck_ana});
    \item the degree to which generated elaborations cohere with the original problem to be solved, i.e., the \emph{verisimilitude}, \emph{pertinence}, and \emph{faithfulness} of the elaborations (Subsection~\ref{subsec:verisim_pert_faithf});
    \item the degree to which generated elaborations syntactically resemble the problem-specific "ideal" elaborations, as alluded to in the fewshot sample solutions (Subsection~\ref{subsec:similsamplesolu});
    \item the degree to which piecemeal elaborations are syntactically redundant and internally coherent (Subsection~\ref{subsec:redundintcoh}). 
\end{enumerate}

\subsection{Do generated elaborations facilitate the application of the simple heuristic?}\label{subsec:luck_ana}

If the model is initially lucky, i.e. there are few effective distractors, its syntactic heuristic is highly effective and adding additional context just tends to reduce overall accuracy (Figure~\ref{fig:1st-o-acc}b). Yet, what's the mechanism underlying the performance boost due to dynamic problem elaboration we have observed? Does problem elaboration (A) block the application of the syntactic heuristic whenever it's not successful and incite the model to deploy a better prediction strategy? Or, (B) does it expand the problem in a way such that the simple syntactic heuristic becomes more effective if applied on the extended context?

\begin{figure}
  \centering
  \begin{tabular}{@{}c@{}c@{}} 
    \includegraphics[height=0.26\linewidth]{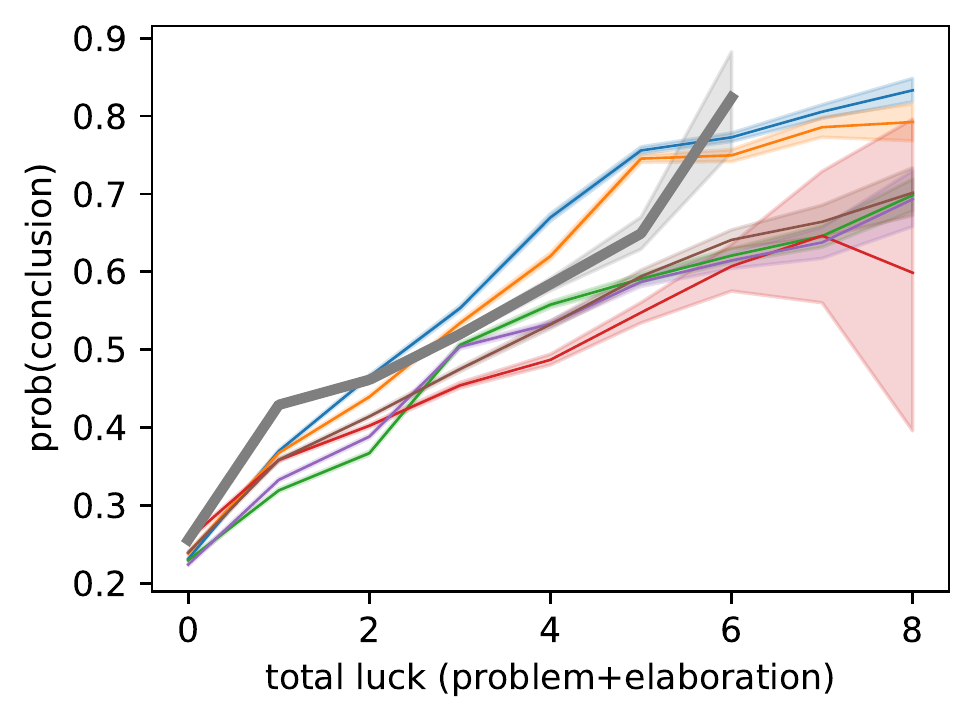} & 
    \includegraphics[height=0.26\linewidth]{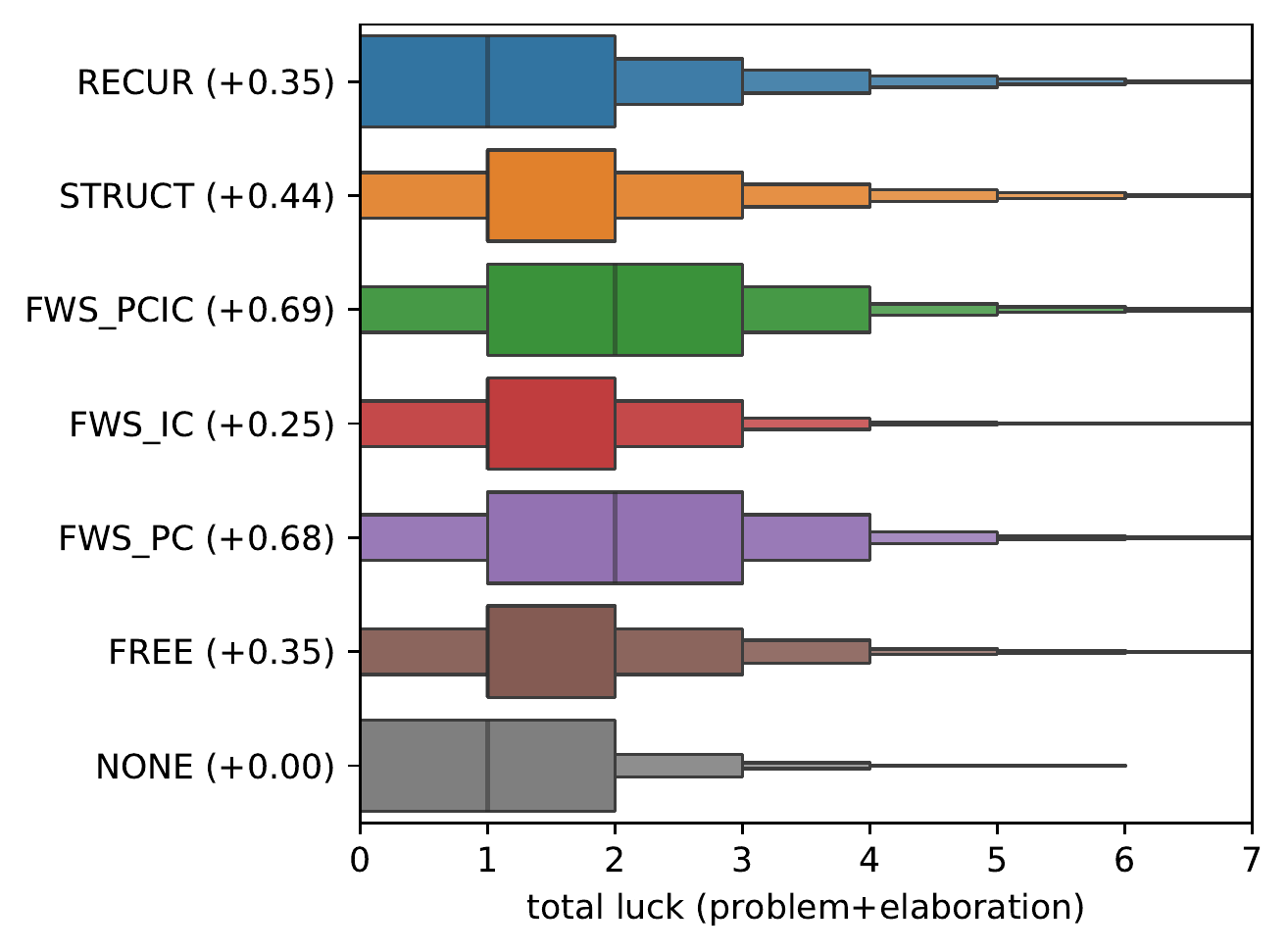} \\
  (a) & (b) \\
\end{tabular}
  \caption{Prediction score of the correct answer (conclusion) and total epistemic luck -- classified according to underlying elaboration type. (a): Mean prediction score in function of epistemic luck, baseline \emph{none} thick, colors as in (b). (b): Distribution of total luck per problem for different types of problem elaboration, mean increase relative to baseline \emph{none} in brackets.}
  \label{fig:total_luck}
\end{figure}

To address these question, we introduce the notion of \emph{total (epistemic) luck} -- a counterpart to the concept of effective distraction (Subsection~\ref{subsec:data}). Total epistemic luck refers to the number of occurrences of the conclusion's predicate both in the original problem description and the generated elaboration (provided the conclusion's predicate is not preceded by "not"). Given a context with high total luck, the simple syntactic heuristic is likely to yield the correct answer to a ChainRuler problem. Figure~\ref{fig:total_luck}a plots the model's prediction score of the correct answer (cf.\ Subsection~\ref{subsec:predicting}) as a function of total epistemic luck for different types of elaboration. For baseline \emph{none} (gray), we observe a clear linear relationship: the model scores the correct answer with p=.25 if total luck equals 0, compared to p>.8 if total luck is 6. This is another way to say that the model uses a simple syntactic heuristic for prediction. Now, importantly, we observe a similar relationship for the predictions based on elaborations, too (where the relationship is slightly stronger for piecemeal elaborations). This suggests that the model is relying on the syntactic heuristic no matter whether it bases its prediction on the original or the dynamically expanded context. What seems to drive the performance boost by problem elaboration, then, is an expansion of the context that facilitates the application of the simple heuristic. In other words, the model is overall more lucky with than without problem elaboration. In fact, and consistent with our analysis, Figure~\ref{fig:total_luck}b shows that problem elaboration increase total epistemic luck by, on average, 0.35-0.7 points.

\subsection{Do generated elaborations cohere with the problem to be solved?}\label{subsec:verisim_pert_faithf}

Verisimilitude, pertinence and faithfulness measure the degree to which an elaboration coheres with different aspects of a given problem.

\begin{small}
  \begin{tabular}{@{}p{0.14\linewidth}@{\rule[0mm]{0.02\linewidth}{0cm}}p{0.41\linewidth}@{\rule[0mm]{0.02\linewidth}{0cm}}p{0.41\linewidth}@{}}
    \toprule
    & \textit{Informal explication} & \textit{Formal operationalization} \\ \midrule
    \textit{Verisimilitude:} & degree to which the elaboration is semantically similar to the ground truth & cosine similarity between sentence-embed-dings of \texttt{elaboration} and \texttt{conclusion}\\
    \textit{Pertinence:} & degree to which the elaboration is semantically similar to the disjunction of possible answers & cosine similarity between sentence-embed-dings of \texttt{elaboration} and \texttt{question}\\
    \textit{Faithfulness:} & degree to which the elaboration is semantically similar to the problem description (premises) & cosine similarity between sentence-embed-dings of \texttt{elaboration} and \texttt{context}\\
    \bottomrule
  \end{tabular}
\end{small}

Transformer embeddings offer an elegant operationalization of the metaphoric notion of semantic similarity. (Technically speaking, we calculate cosine similarity between the DistilBERT-embeddings of the corresponding texts \citep{reimers-2019-sentence-bert}.)

\begin{figure}
  \centering
  \begin{tabular}{@{}c@{}c@{}c@{}} 
  \small\textit{(a) Verisimilitude} & \small\textit{(b) Pertinence} & \small\textit{(c) Faithfulness}\\
  \includegraphics[height=0.28\linewidth]{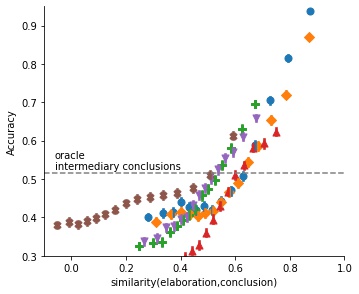} & 
  \includegraphics[height=0.28\linewidth]{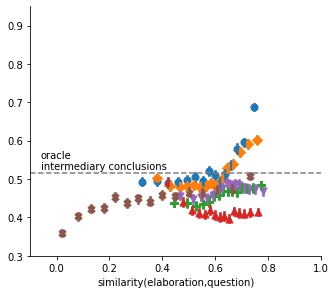} & 
  \includegraphics[height=0.28\linewidth]{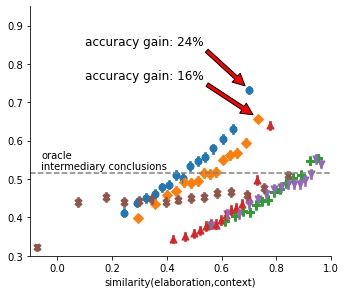} \\
  \includegraphics[height=0.28\linewidth]{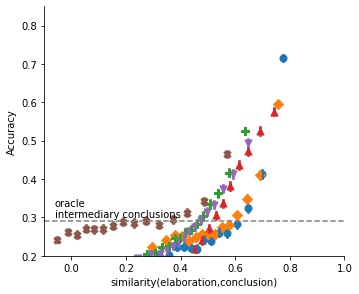} &
  \includegraphics[height=0.28\linewidth]{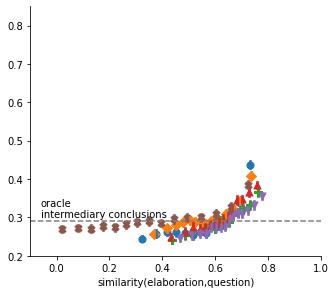} &
  \includegraphics[height=0.28\linewidth]{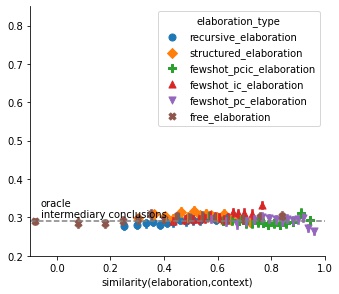} \\ 
  \end{tabular}
  \caption{Accuracy in ChainRuler task for six types of elaborations as a function of (a) their verisimilitude, that is the semantic similarity between generated elaboration and correct answer (conclusion), (b) their pertinence, that is the semantic similarity between generated elaboration and sequence of possible answers, and (c) their faithfulness, that is the semantic similarity between generated elaboration and context. Top row: without contraposition. Bottom row: with contraposition.}
  \label{fig:similarity}
\end{figure}

Figure~\ref{fig:similarity}a plots GPT-2's accuracy on ChainRuler tasks as a function of the elaborations' verisimilitude. As expected, the more a dynamically generated elaboration resembles the correct answer, the more likely the model is to provide the correct answer given the elaboration. This observation is consistent with our analysis of total epistemic luck (Subsection~\ref{subsec:luck_ana}) as well as with the finding that oracle elaborations which just repeat the correct answer (maximum verisimilitude) boost accuracy levels above 80\% (cf.\ Appendix~\ref{sec:app_elabmethods}). Moreover, these highly plausible results also corroborate the method of semantic similarity analysis based on transformer embeddings.

Figure~\ref{fig:similarity}b plots GPT-2's accuracy on ChainRuler tasks as a function of the the generated elaboration's semantic similarity to the problem's question, which presents the three alternative answers. We observe a positive relation between pertinence and accuracy, especially for recursive and structured elaborations. If a piecemeal elaboration really addresses the question, then its, on average, more likely to be effective. 

Figure~\ref{fig:similarity}c plots GPT-2's accuracy on ChainRuler tasks as a function of the generated elaboration's faithfulness to the problem description. For ChainRuler tasks without contraposition, we obtain clear and highly plausible results (upper row): The more faithful the dynamic elaboration, the more effective it is in terms of helping the model to predict the correct answer. The relationship is most pronounced for piecemeal elaborations, such that the most faithful (top 7\%) recursive and structured elaborations increase accuracy by 24 respectively 16 percentage points (as compared to no elaboration). Concerning the ChainRuler tasks with contraposition (bottom plot in Figure~\ref{fig:similarity}c), faithfulness as measured by embedding similarity seems to have no effect on accuracy. However, a manual re-analysis of the data reveals that faithfulness is positively correlated with accuracy and that cosine similarity between BERT-embeddings simply fails to reflect deductive implications as soon as contraposition is involved \citep[see also][]{kassner2020negated}. 

All in all, variance in elaboration effectiveness can partly be explained in terms of coherence with the original problem (as confirmed by a logistic regression analysis: the $R^2$ statistics with \emph{depth} and \emph{effective distraction} as endogenous variables equals 2.7\%, but increases to 9.8\% if verisimilitude, pertinence and faithfulness are included as further explanatory variables). A further take-away from Figure~\ref{fig:similarity} is that piecemeal elaborations benefit most from cohering with the problem. The effectiveness of free and fewshot elaborations increases to a much lesser extent with rising pertinence or faithfulness. This might be due to the following difference: Free and fewshot elaborations may resemble a question or a problem description in argumentatively irrelevant ways (simply by repeating the question or mimicking the syntactic structure of the rules). Piecemeal elaborations, however, consist by design in statements about the problem's subject and are hence much more likely to cohere with the problem in inferentially relevant ways, if they cohere with it at all.

\subsection{Do generated elaborations resemble ideal elaborations?}\label{subsec:similsamplesolu}

We consider two kinds of problem specific "ideal" elaborations. Given a ChainRuler problem, the \emph{perfect proof chain} consists in the \texttt{fact}, the \texttt{rule chain} (in correct order), and the final \texttt{conclusion}. The \emph{intermediary and final conclusions} simply are the intermediary conclusions (in the order they can be inferred by applying the rules) plus the final \texttt{conclusion}. We use BLEU2-scores to measure the extent to which a given problem elaboration syntactically resembles the corresponding ideal elaboration.

\begin{figure}
  \centering
  \begin{tabular}{@{}r@{}r@{}r@{}r@{}}
  \parbox[b]{0.33\linewidth}{\centering \small \textit{(a) Syntactic similarity to perfect proof chain}}  &
  \parbox[b]{0.33\linewidth}{\centering \small \textit{(b) Syntactic similarity to intermediary and final conclusions}}  &
  \parbox[b]{0.12\linewidth}{\centering \small \textit{(c) Internal redundancy}}  &
  \parbox[b]{0.12\linewidth}{\centering \small \textit{(d) Internal coherence}}  \\
  \includegraphics[height=15.2mm]{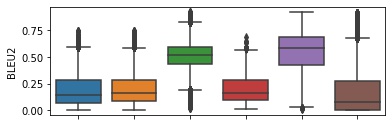} & 
  \includegraphics[height=15.3mm]{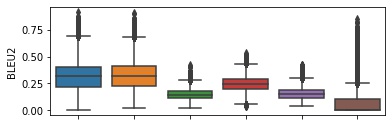} & 
  \includegraphics[height=15.6mm]{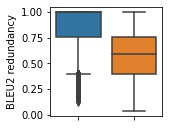} & 
  \includegraphics[height=15.5mm]{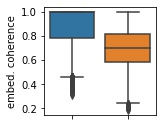} \\ 
  \includegraphics[height=19mm]{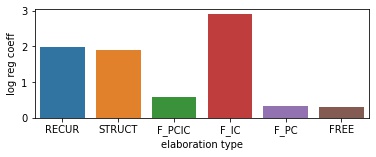} & 
  \includegraphics[height=19mm]{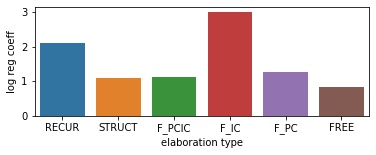} & 
  \includegraphics[height=19mm]{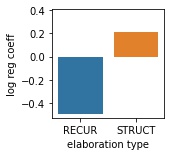} & 
  \includegraphics[height=19mm]{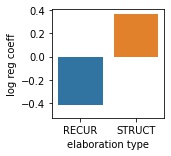} \\ 
  \end{tabular}
  \caption{Distributions of similarity values (top row) and logistic regression coefficients (bottom row). For each column (a)--(d) and elaboration type, a logistic regression with accuracy as exogenous variable and depth, breadth, and corresponding elaboration similarity as endogenous variables is carried out.}
  \label{fig:similarity3}
\end{figure}

As the boxplot in Figure~\ref{fig:similarity3}a reveals, fewshot PC and fewshot PCIC elaborations are syntactically highly similar to the corresponding perfect proof chains. Similarly, fewshot IC elaborations syntactically resemble intermediary and final conclusions to a greater extent than the other free and fewshot elaborations -- albeit less so than piecemeal elaborations (Figure~\ref{fig:similarity3}b). Thus, the fewshot samples are clearly shaping the generated elaborations. Yet, does this effect pay off in terms of accuracy? Do elaborations which syntactically resemble an "ideal" elaboration tend to be more effective? The barplots in Figure~\ref{fig:similarity3}a and b answer this question, by reporting the degree to which syntactic similarity leads to higher accuracy, as measured by a logistic regression (which controls for depth and breadth). Fewshot IC and piecemeal elaborations tend to be much more effective if they resemble the perfect proof chain. Accordingly, one reason for the overall poor performance of fewshot IC (cf.\ Appendix~\ref{sec:app_elabmethods}) seems to be that the model mostly fails to generate the correct intermediary and final conclusions, even if "told so" by fewshot examples. This is not the case at all for fewshot PC and fewshot PCIC elaborations. As soon as the model is "told to" generate proof chains, syntactic similarity to the ideal proof chain ceases to be an indicator of accuracy.

\begin{table}
  \caption{Relative accuracy on ChainRuler tasks with unnegated \texttt{fact}, displayed as difference to absolute accuracy values averaged over all tasks.}
  \label{table:accuracy-unnegated-facts}
  \centering
  \begin{footnotesize}
  \begin{tabularx}{\linewidth}{l *{11}{Y}} 
    \toprule
    & \multicolumn{6}{c}{\emph{Elaborations}} & \multicolumn{3}{c}{\emph{Baselines}} & \multicolumn{2}{c}{\emph{Oracles}}                   \\
    \cmidrule(r){2-7} \cmidrule(r){8-10} \cmidrule(r){11-12}
    \emph{cntrp}   & \tiny{FREE} & \tiny{F\_IC} & \tiny{F\_PC} & \tiny{F\_PCIC} & \tiny{STRUCT} & \tiny{RECUR} & \tiny{NONE}  & \tiny{ANSWS} & \tiny{CONTXT} & \tiny{INTERM} & \tiny{FINAL} \\ \midrule
    False & \cellcolor{blue!9}0.9 &  \cellcolor{blue!24}2.4 & \cellcolor{blue!50}5.0 & \cellcolor{blue!44}4.4 & \cellcolor{blue!17}1.7 & \cellcolor{blue!39}3.9 & \cellcolor{blue!12}1.2  & \cellcolor{blue!4}0.4 & \cellcolor{blue!30}3.0 & \cellcolor{blue!35}3.5 & \cellcolor{blue!30}3.0 \\
    True &  \cellcolor{red!23}-2.3 & \cellcolor{blue!7}0.7 & \cellcolor{red!11}-1.1 & \cellcolor{red!8}-0.8 & \cellcolor{red!17}-1.7 & \cellcolor{blue!2}0.2 &  \cellcolor{red!27}-2.7  & \cellcolor{red!5}-0.5 & \cellcolor{red!29}-2.9 & \cellcolor{red!25}-2.5 & \cellcolor{red!10}-0.1 \\
    \bottomrule
  \end{tabularx} 
  \end{footnotesize}
\end{table}

The ability of GPT-2 to generate and exploit ideal elaborations -- in particular: proof chains -- is strongly influenced by the presence of negations in the problem description. Once more, "not" turns out to be a trouble-maker. To see this, we consider ChainRuler tasks whose singular premise (\texttt{fact}) is not negated. Table~\ref{table:accuracy-unnegated-facts} reports the accuracy difference in these tasks as compared to all tasks. The fewshot elaborations with proof chain examples are significantly more effective with unnegated \texttt{facts} (actually, fewshot PC now outperforms baseline none). This seems to be not only due to the generation of more accurate proof chains, but also to a better ability of the model to tap on good elaborations, as the increased accuracy of oracle elaborations suggests.

\subsection{Are piecemeal elaborations syntactically redundant and internally coherent?}\label{subsec:redundintcoh}

Piecemeal elaborations are composed of four separately generated statements about the problem's \texttt{subject}. We assess the \textit{syntactic internal redundancy} of such an elaboration by averaging over pairwise BLEU2-scores, and take the mean cosine similarity of the sentence-embeddings to be a measure of the elaboration's \textit{semantic internal coherence}.

As Figure~\ref{fig:similarity3}c,d shows, piecemeal elaborations are highly redundant and internally coherent; recursive elaborations even more so than structured ones. (Roughly half of the recursive elaborations simply consist in  one and the same sentence repeated four times.) Redundancy/coherence has, however, opposite effects on the effectiveness of recursive versus structured elaborations. Recursive elaborations are the more effective, the less syntactically redundant / semantically coherent they are. Structured elaborations, in contrast, gain from redundancy and coherence. 

These findings can be explained in terms of the underlying generation methods. With \textit{recursive elaboration}, a first sentence about the \texttt{subject} is generated an appended to the context. Then, the model is prompted to generate a second statement about the \texttt{subject} given the updated context: Either the model "sees" what else can be inferred about \texttt{subject} given the newly added first sentence, and generates another sentence -- which leads to a sensible proof chain, and to low redundancy. Or the model does not "see" what else can be inferred from the updated context, and then simply generates again what it has generated before (additionally encouraged to do so by positive feedback effects observed in \citet{holtzman2019curious}), namely the first sentence -- which is a sign of poor inferential insight and results in high redundancy. That's why low redundancy goes along with high effectiveness of recursive elaborations. Now, the four individual sentences that make up a \textit{structured elaboration} are generated independently of each other. So, the more confident the model is about how to complete a sentence about the \texttt{subject}, the more likely it is that this sentence will be generated several times when prompting the model independently of each other. For structured elaboration, redundancy and internal coherence are therefore indicators of confidence, which explains -- assuming that models are all in all decently calibrated for ChainRuler tasks -- why high redundancy coincides with high accuracy.

\section{Conclusion and Future Work}\label{sec:discussion}

In this paper, we introduce ChainRuler, a dataset for multi-hop deductive argumentation, and assess GPT-2's zero-shot ability both to solve the inference tasks and to generate \emph{effective problem elaborations}, i.e., texts which -- once added to the context -- improve performance. Our main findings are:

\begin{itemize}
    \item GPT-2 follows a simple heuristic when prompted with a ChainRuler reasoning problem -- which leads to high accuracy in benevolent cases, but causes systematic bias as soon as \emph{effective distraction} is high or the task involves \textit{contraposition}: pre-trained GPT-2 then performs much worse than random guessing. (Section~\ref{sec:results})     
    \item Dynamic context expansion with generated problem elaborations can, depending on the problem characteristics, increase accuracy by up to 9\%, i.e., by an order of magnitude observed in comparable experiments yet other tasks \citep{shwartz2020unsupervised,saha2020prover}. Elaborations possess, depending on how they are generated, characteristic "accuracy fingerprints" over the problem spectrum. (Section~\ref{sec:results})
    \item Dynamic problem elaboration doesn't prevent the model from applying its heuristic. On the contrary, it expands the context so that the syntactic heuristic can be applied more successfully. Bluntly put: The reasoning is all in the context generation, the final prediction remains "stupid". (Subsection~\ref{subsec:luck_ana})
    \item Variance in elaboration effectiveness can be explained in view of the extent to which an elaboration coheres with the problem to be solved. Moreover, the most faithful so-called recursive and structured elaborations boost accuracy by 24\% resp.\ 16\%, compared to the no elaboration treatment. (Subsection~\ref{subsec:verisim_pert_faithf})  
    \item Fewshot learning (in the sense of \citep{brown2020language}) powerfully shapes the generated elaborations (Subsection~\ref{subsec:similsamplesolu}), but does not lead to significantly stronger overall performance (Section~\ref{sec:results}). Rather, different types of fewshot elaborations excel under different kinds of problem characteristics (Section~\ref{sec:results}) -- especially so when negations are absent in the corresponding problem descriptions (Subsection~\ref{subsec:similsamplesolu}).
    \item Redundancy is not necessarily a flaw of an elaboration. Rather, repeating a statement over and again can be a sign of a model's strong confidence and enable it to successfully exploit the generated elaboration (Subsection~\ref{subsec:redundintcoh}).  
\end{itemize}

All these results are obtained with pre-trained GPT-2 and without further fine-tuning. This is certainly the reason for why we observe substantial, yet still clearly limited inference skill and ability to generate effective problem elaborations. This said, it seems worthwhile to explore, in future research, whether generative Transformer language models can \textbf{learn to think aloud}. Obviously, there are alternative set-ups for training language models to generate and exploit sound problem elaborations, for example:

\begin{itemize}
    \item The language model is fine-tuned on the specific task, e.g., the ChainRuler data.
    \item The language model is fine-tuned on a given corpus of good problem elaborations (like the ones considered in Subsection~\ref{subsec:similsamplesolu}).
    \item The language model is fine-tuned on a dynamically evolving dataset: The model generates free elaborations. Those elaborations that increase prediction accuracy to the greatest extent are added to the training data. The model is fine tuned on the training data. Next, another round of free elaborations are generated; once more, the best elaborations are added to the training data, and so on.
\end{itemize}

Besides improvements in accuracy and reliability, transfer learning effects would be of major interest in this context. For instance, it would be important to study whether language models are able to generalize a problem solving heuristic, and to produce effective elaborations beyond the tasks they have been trained on.

\subsubsection*{Acknowledgements}

We would like to thank the members of the Aristo group at Allen AI for valuable feedback on earlier versions of this work.  

\appendix

\section{Sample solutions used in fewshot elaborations}

\begin{description}
    \item[Fewshot IC] 
    Here is what we know: If someone is lonely, then they are not brown. If someone is big, then they are not popular. Bill is brown. Does this mean that Bill is lonely, is not lonely, or is not popular? Explain! Well, it says that Bill is brown. It follows that Bill is not lonely. Therefore, Bill is not lonely. \newline Here is what we know: If someone is boring, then they are tall. If someone is gray, then they are clever. If someone is clever, then they are tall. Chris is gray. If someone is tired, then they are small. Does this mean that Chris is tall, is not tall, or is small? Explain! Well, it says that Chris is gray. It follows that Chris is clever. And therefore, Chris is tall.
    \item[Fewshot PC]
    Here is what we know: If someone is lonely, then they are not brown. If someone is big, then they are not popular. Bill is brown. Does this mean that Bill is lonely, is not lonely, or is not popular? Explain! Well, it says that Bill is brown. If someone is brown, then they are not lonely. Therefore, Bill is not lonely. \newline Here is what we know: If someone is boring, then they are tall. If someone is gray, then they are clever. If someone is clever, then they are tall. Chris is gray. If someone is tired, then they are small. Does this mean that Chris is tall, is not tall, or is small? Explain! Well, it says that Chris is gray. If someone is gray, then they are clever. If someone is clever, then they are tall. Therefore, Chris is tall.
    \item[Fewshot PCIC]
    Here is what we know: If someone is lonely, then they are not brown. If someone is big, then they are not popular. Bill is brown. Does this mean that Bill is lonely, is not lonely, or is not popular? Explain! Well, it says that Bill is brown. If someone is brown, then they are not lonely. Therefore, Bill is not lonely. \newline Here is what we know: If someone is boring, then they are tall. If someone is gray, then they are clever. If someone is clever, then they are tall. Chris is gray. If someone is tired, then they are small. Does this mean that Chris is tall, is not tall, or is small? Explain! Well, it says that Chris is gray. If someone is gray, then they are clever. It follows that Chris is clever. If someone is clever, then they are tall. And therefore, Chris is tall.    
\end{description}

\section{Effectiveness of Elaboration Methods}\label{sec:app_elabmethods}

\begin{figure}
  \centering\footnotesize
  \begin{tabular}{@{}c@{}c@{}c@{}c@{}c@{}c@{}}
    \includegraphics[height=25mm]{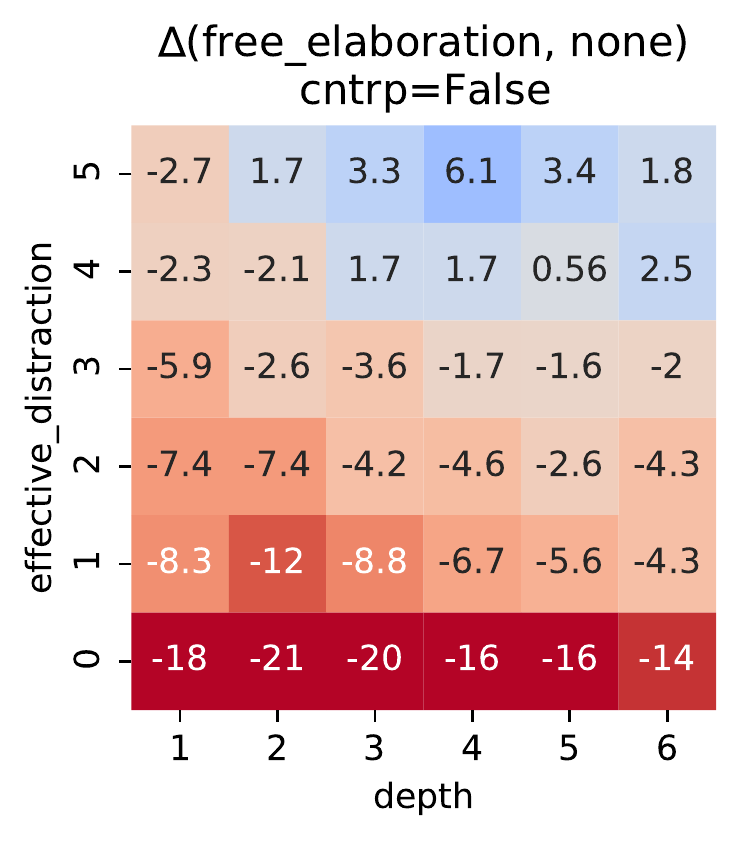} &  
    \includegraphics[height=25mm]{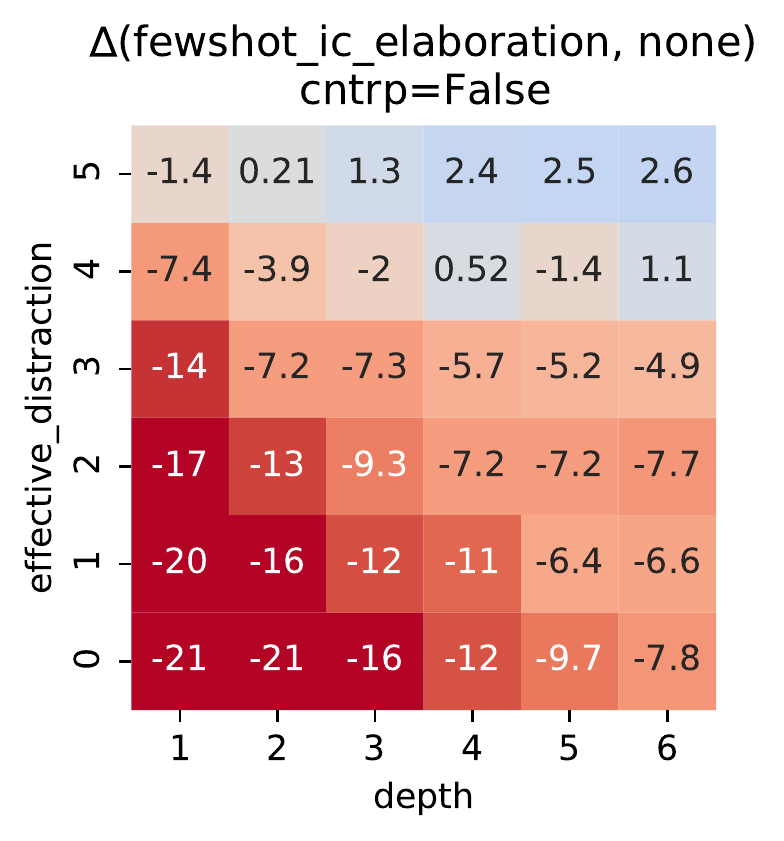}& 
    \includegraphics[height=25mm]{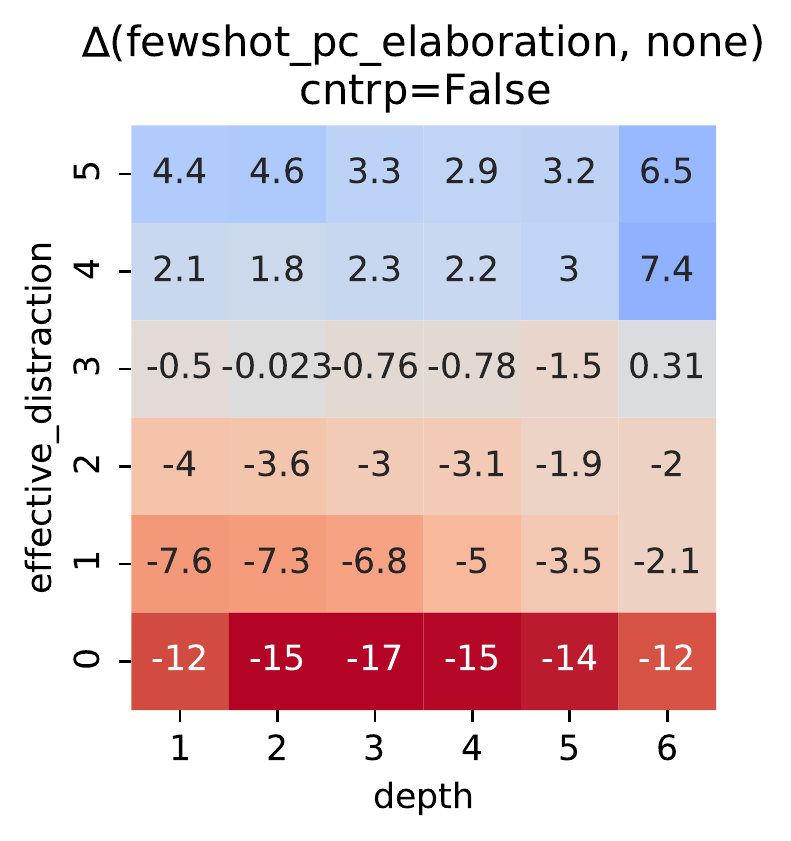} &
    \includegraphics[height=25mm]{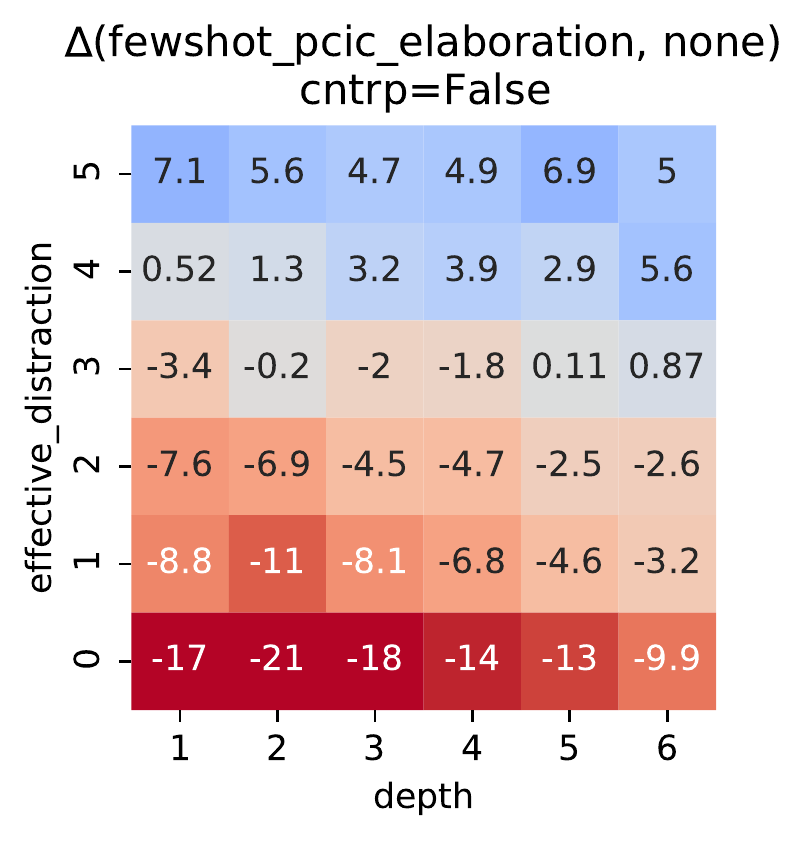} &
    \includegraphics[height=25mm]{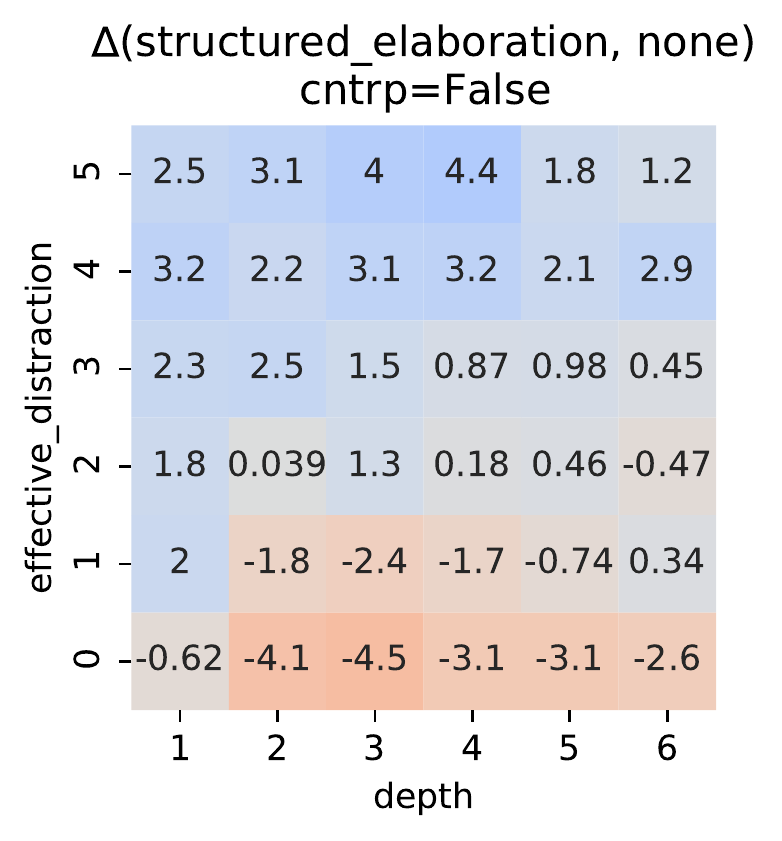}&
    \includegraphics[height=25mm]{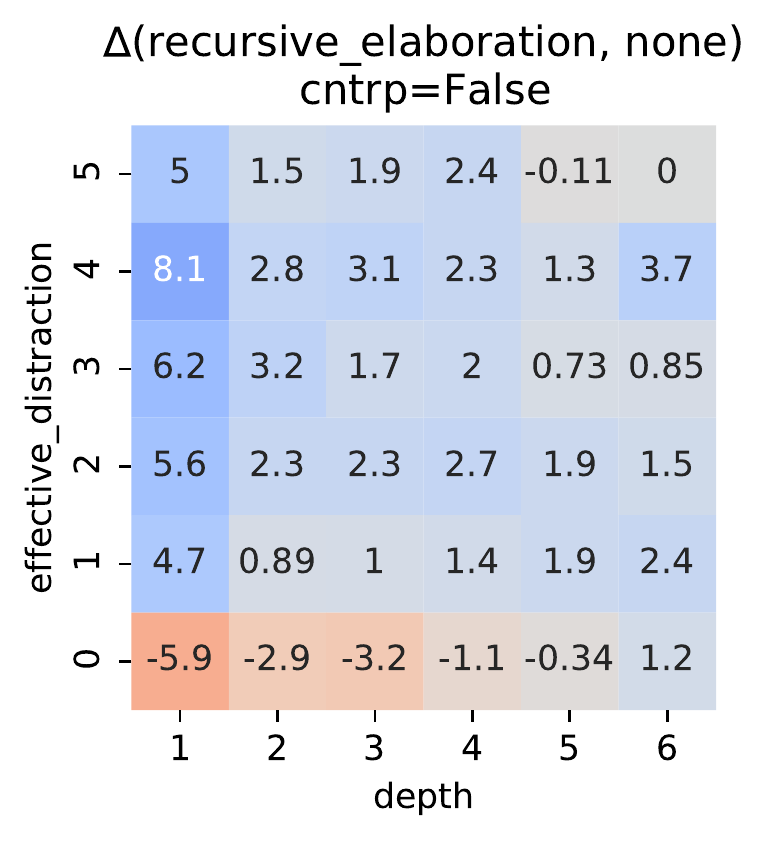}\\
    \includegraphics[height=25mm]{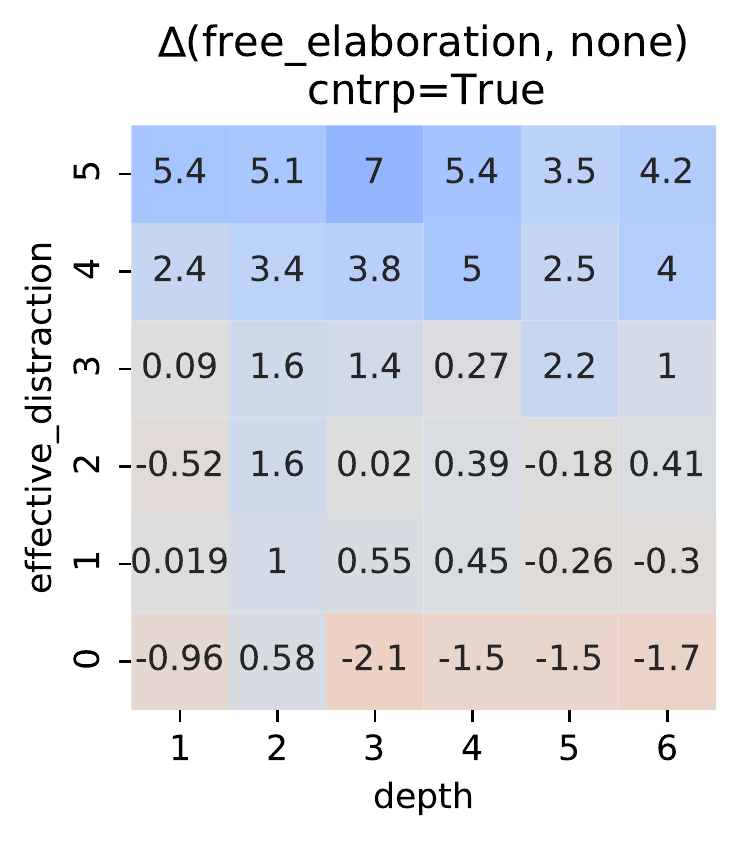} &  
    \includegraphics[height=25mm]{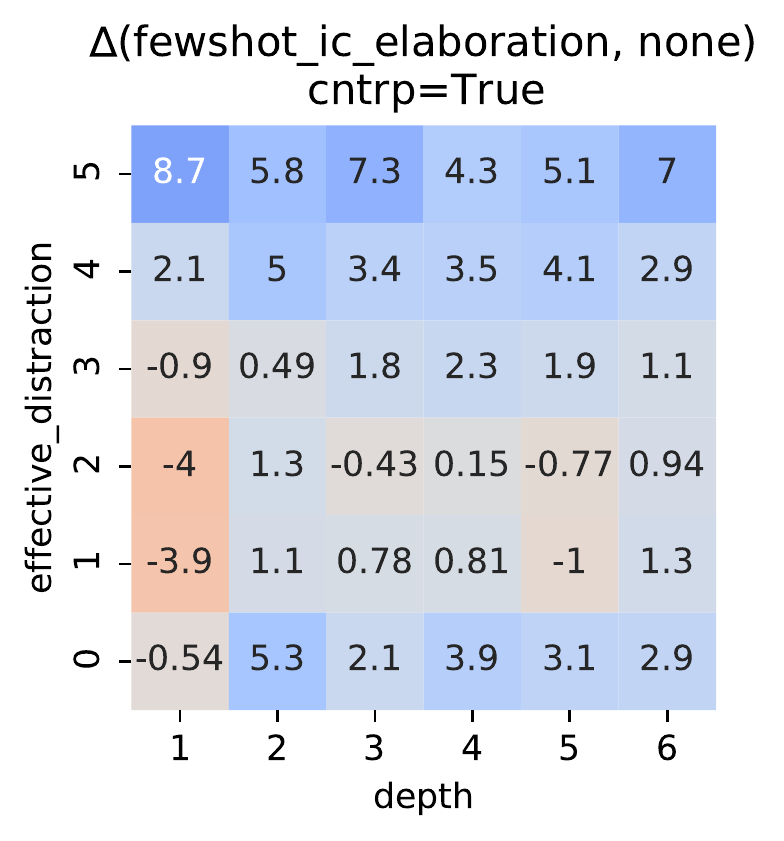}& 
    \includegraphics[height=25mm]{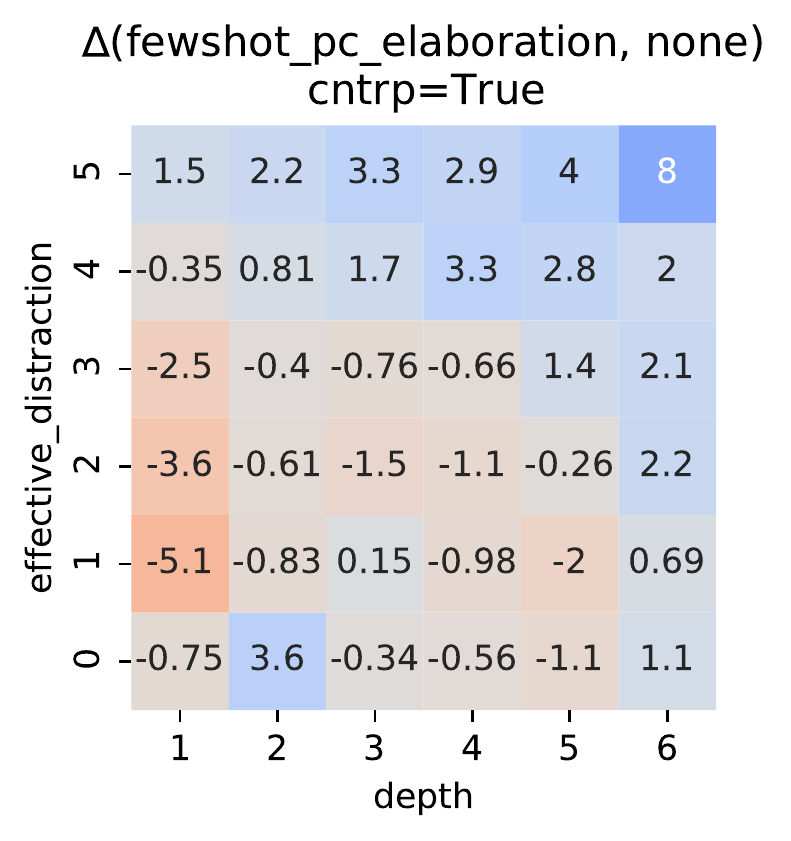} &
    \includegraphics[height=25mm]{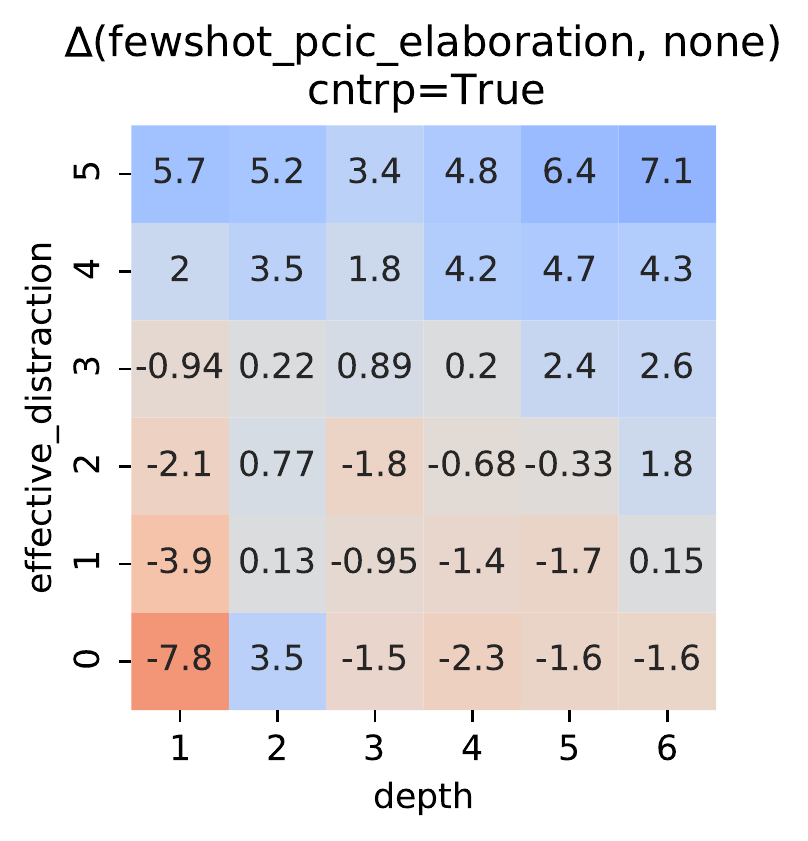} &
    \includegraphics[height=25mm]{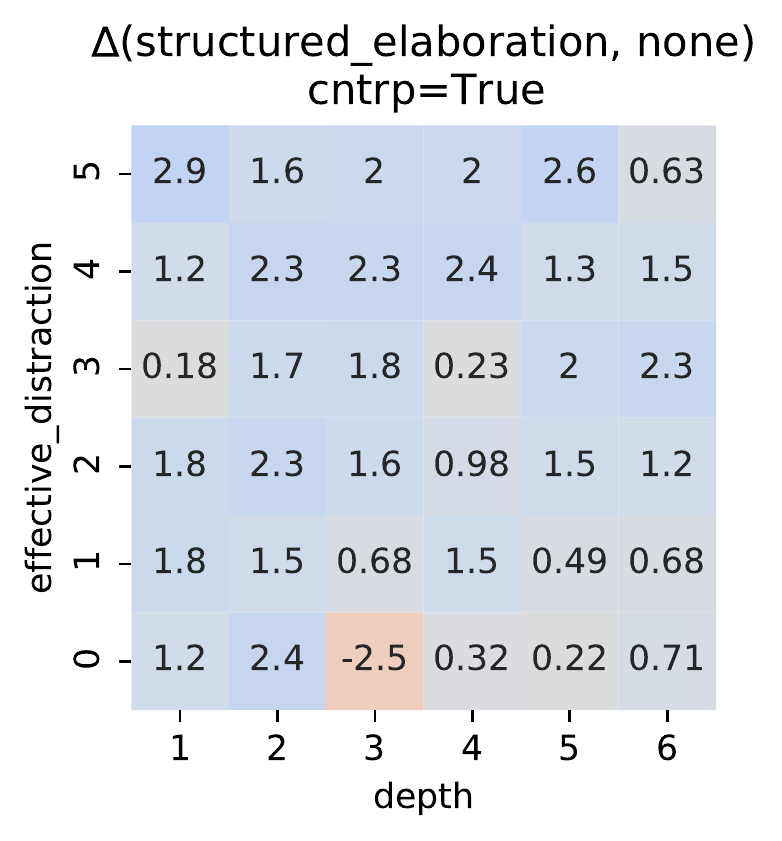}&
    \includegraphics[height=25mm]{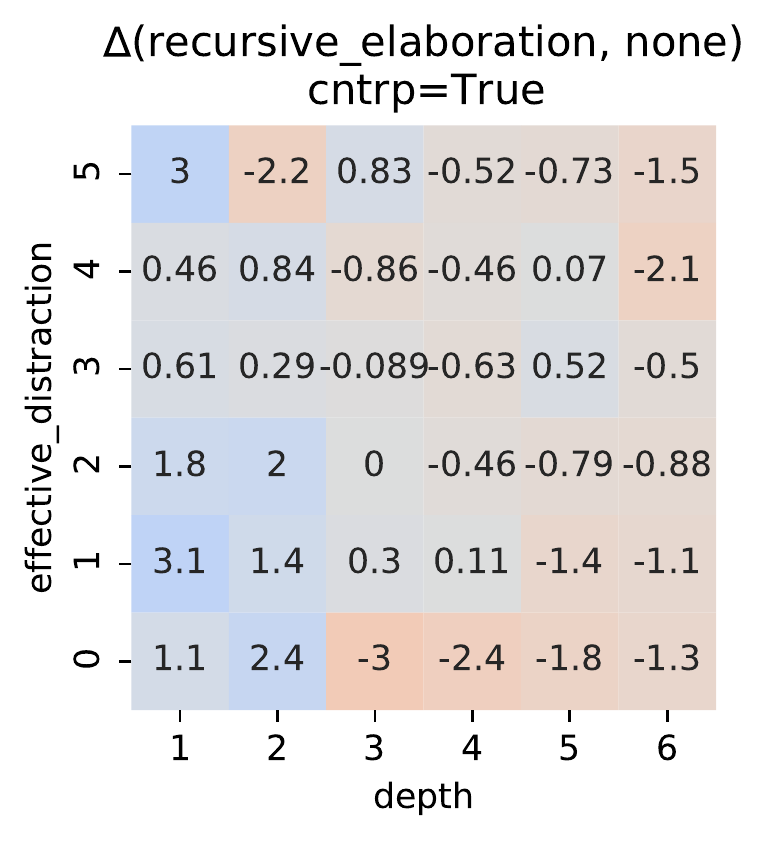}\\
    \includegraphics[height=25mm]{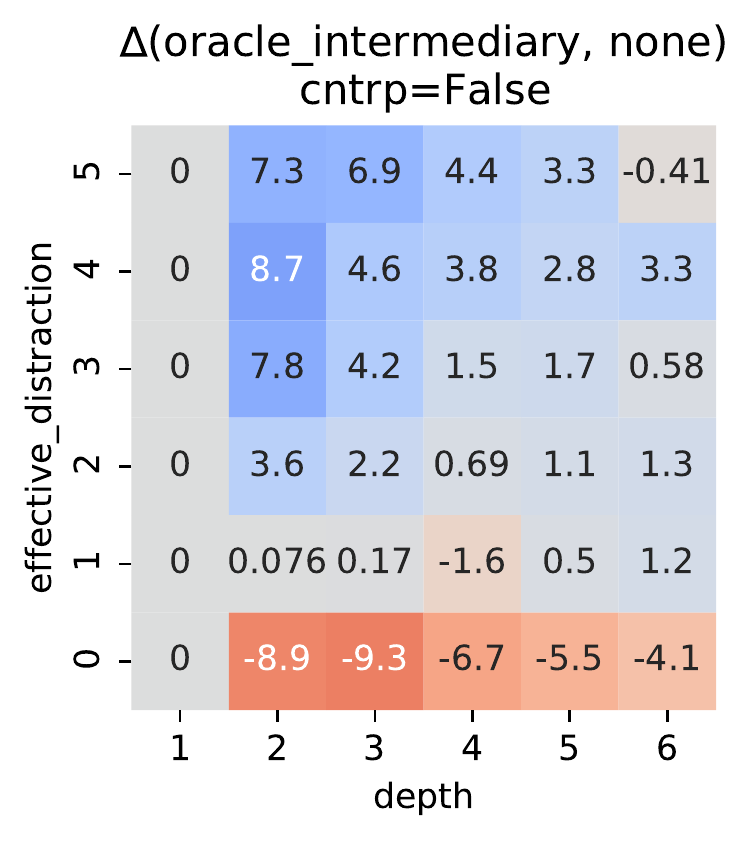} &
    \includegraphics[height=25mm]{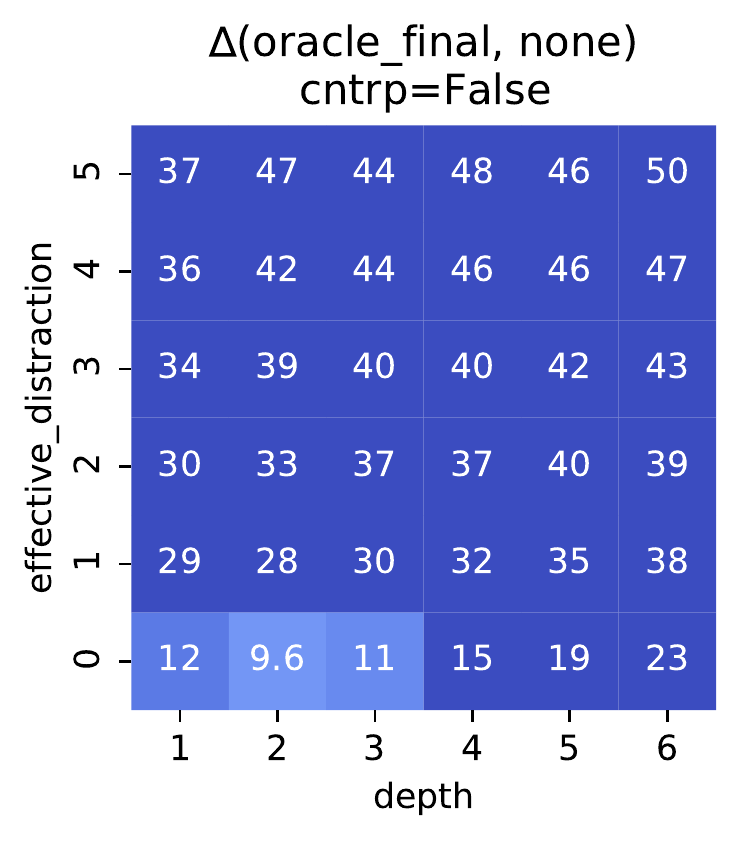} &
    \includegraphics[height=25mm]{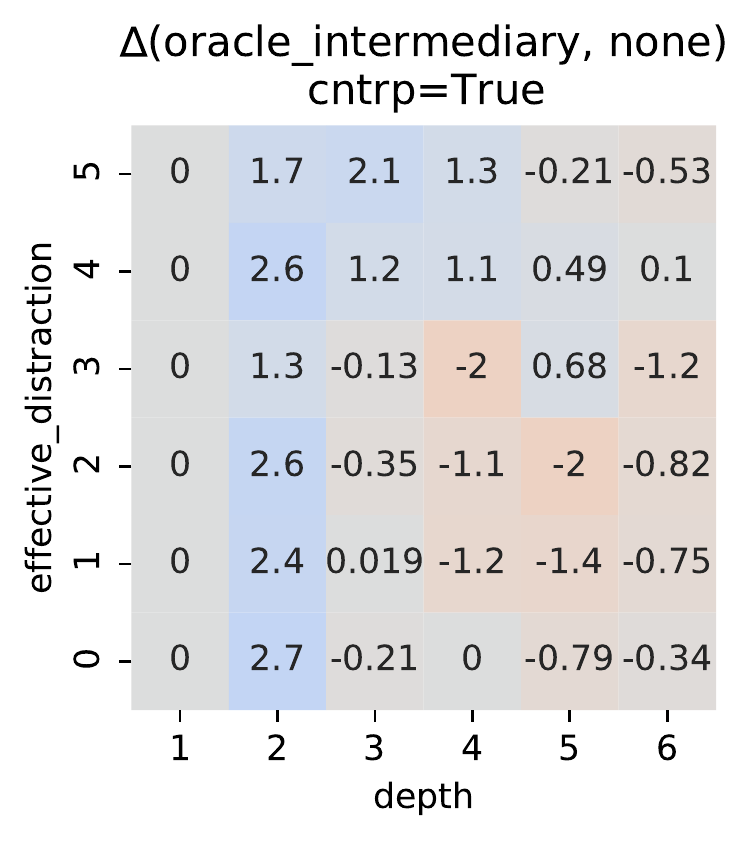} &
    \includegraphics[height=25mm]{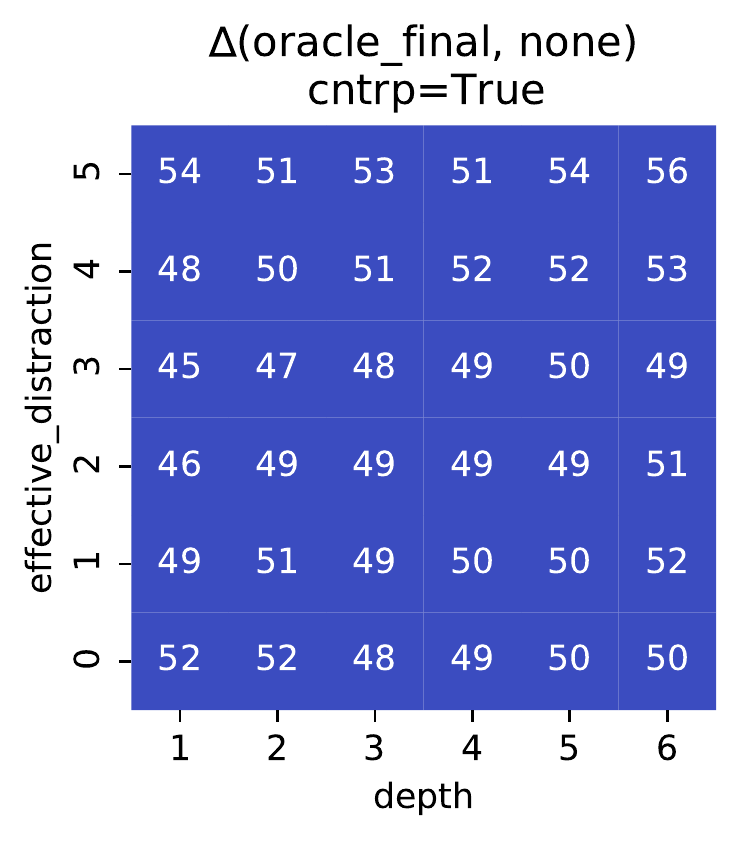}\\
  \end{tabular}
  
  \caption{Effectiveness of elaborations methods.}
  \label{fig:1st-o-acc_detailed}
\end{figure}

Figure~\ref{fig:1st-o-acc_detailed} reports detailed accuracy gains achieved by various kinds of dynamic problem elaborations, including oracles.

\bibliographystyle{plainnat}
\bibliography{bib_all.bib}

\end{document}